\documentclass[lettersize,journal]{IEEEtran}
\usepackage{amsmath,amsfonts}
\usepackage{algorithmic}
\usepackage{array}
\usepackage[caption=false,font=normalsize,labelfont=sf,textfont=sf]{subfig}
\usepackage{textcomp}
\usepackage{stfloats}
\usepackage{url}
\usepackage{verbatim}
\usepackage{graphicx}
\usepackage{amsmath}
\usepackage{amssymb}
\usepackage[dvipsnames,table,xcdraw]{xcolor}
\usepackage{textcomp}
\usepackage{verbatim}
\usepackage{cite}
\usepackage{booktabs}
\usepackage{tabularx}
\usepackage{amsfonts}
\usepackage{multirow}
\usepackage{bbding}
\usepackage{array}
\usepackage{pifont}
\usepackage{colortbl}
\usepackage{wasysym}
\usepackage{times}  % DO NOT CHANGE THIS
\usepackage{helvet}  % DO NOT CHANGE THIS
\usepackage{courier}  % DO NOT CHANGE THIS
\usepackage{graphicx} % DO NOT CHANGE THIS
\usepackage{caption} % DO NOT CHANGE THIS AND DO NOT ADD ANY OPTIONS TO IT

\usepackage{algorithm}
\usepackage{algorithmic}

\usepackage{amsmath}
\usepackage{amssymb}
\usepackage{multirow}
\usepackage{xcolor}
\usepackage{newfloat}
\usepackage{listings}
\usepackage{pifont}
\usepackage{colortbl}
\definecolor{mygray}{gray}{.92}

\def\BibTeX{{\rm B\kern-.05em{\sc i\kern-.025em b}\kern-.08em
    T\kern-.1667em\lower.7ex\hbox{E}\kern-.125emX}}

\newlength\savewidth

\def\BibTeX{{\rm B\kern-.05em{\sc i\kern-.025em b}\kern-.08em
    T\kern-.1667em\lower.7ex\hbox{E}\kern-.125emX}}
\usepackage{balance}

\DeclareRobustCommand*{\IEEEauthorrefmark}[1]{%
    \raisebox{0pt}[0pt][0pt]{\textsuperscript{\footnotesize\ensuremath{#1}}}}

\makeatletter
\newcommand{\thickhline}{%
    \noalign {\ifnum 0=`}\fi \hrule height 0.8pt
    \futurelet \reserved@a \@xhline
}

\begin{document}

%\author{Jiawei Mao\textsuperscript{\emph{a,b},\textdagger}}[orcid=0000-0002-6741-7033]
%\author{Rui Xu\textsuperscript{\emph{a,b},\textdagger}}[orcid=0000-0002-9392-9497]
%\author{Xuesong Yin\textsuperscript{\emph{a,b,}}}[orcid=0000-0001-8455-0840]
%\cormark[1]
%\author{Yuanqi Chang\textsuperscript{\emph{a,b}}}[orcid=0000-0002-9392-9497]
%\author{Binling Nie\textsuperscript{\emph{a,b}}}[orcid=0000-0002-9392-9427]
%\author{AIbin Huang\textsuperscript{\emph{a,b}}}[orcid=0000-0002-9392-9497]
%\address[author1]{School of Media and Design, Hangzhou Dianzi University, Hangzhou, 310018, China}
%\address[author2]{Wenzhou Institute of Hangzhou Dianzi University, Wenzhou, 325038, China}
%\cortext{\textdagger Equal contribution.}
%\cortext[cor1]{Corresponding author.}

\author{
\IEEEauthorblockN{
Jiawei Mao\IEEEauthorrefmark{1,2},
Guangyi Zhao\IEEEauthorrefmark{1,2},
Xuesong Yin\IEEEauthorrefmark{1,2},
Yuanqi Chang\IEEEauthorrefmark{1,2}}

\IEEEauthorblockA{\IEEEauthorrefmark{1}School of Media and Design, Hangzhou Dianzi University, Hangzhou, 310018, China}

\IEEEauthorblockA{\IEEEauthorrefmark{2}Wenzhou Institute of Hangzhou Dianzi University, Wenzhou, 325038, China}

\IEEEauthorblockA{Corresponding Author: Xuesong Yin \quad Email: yinxs@hdu.edu.cn}}

\title{SwinStyleformer is a favorable choice for image inversion}
% \author{IEEE Publication Technology Department
% \thanks{Manuscript created October, 2020; This work was developed by the IEEE Publication Technology Department. This work is distributed under the \LaTeX \ Project Public License (LPPL) ( http://www.latex-project.org/ ) version 1.3. A copy of the LPPL, version 1.3, is included in the base \LaTeX \ documentation of all distributions of \LaTeX \ released 2003/12/01 or later. The opinions expressed here are entirely that of the author. No warranty is expressed or implied. User assumes all risk.}}

\markboth{Journal of \LaTeX\ Class Files,~Vol.~18, No.~9, September~2020}%
{How to Use the IEEEtran \LaTeX \ Templates}

\maketitle

\begin{abstract}
    This paper proposes the first pure Transformer structure inversion network called SwinStyleformer, which can compensate for the shortcomings of the CNNs inversion framework by handling long-range dependencies and learning the global structure of objects. Experiments found that the inversion network with the Transformer backbone could not successfully invert the image. The above phenomena arise from the differences between CNNs and Transformers, such as the self-attention weights favoring image structure ignoring image details compared to convolution, the lack of multi-scale properties of Transformer, and the distribution differences between the latent code extracted by the Transformer and the StyleGAN style vector. To address these differences, we employ the Swin Transformer with a smaller window size as the backbone of the SwinStyleformer to enhance the local detail of the inversion image. Meanwhile, we design a Transformer block based on learnable queries. Compared to the self-attention transformer block, the Transformer block based on learnable queries provides greater adaptability and flexibility, enabling the model to update the attention weights according to specific tasks. Thus, the inversion focus is not limited to the image structure. To further introduce multi-scale properties, we design multi-scale connections in the extraction of feature maps. Multi-scale connections allow the model to gain a comprehensive understanding of the image to avoid loss of detail due to global modeling. Moreover, we propose an inversion discriminator and distribution alignment loss to minimize the distribution differences. Based on the above designs, our SwinStyleformer successfully solves the Transformer's inversion failure issue and demonstrates SOTA performance in image inversion and several related vision tasks.
\end{abstract}

\begin{IEEEkeywords}
Generative adversarial networks, Transformer, Image inversion.
\end{IEEEkeywords}

\section{Introduction}
  
Generative adversarial networks (GANs) \cite{goodfellow2014generative} can generate realistic images. In recent years, GANs \cite{liao2019dr,li2022uphdr,wang2021tms,zhao2020scgan} have played an important role in several visual tasks. Several studies \cite{goetschalckx2019ganalyze,jahanian2019steerability,shen2020interpreting,yang2021semantic,zhou2022hrinversion} have found that the GAN latent space contains rich interpretable semantic information, 
which allows one to use the GAN latent space for several image manipulations \cite{shen2021closed,gu2020image,shen2020interfacegan,zhu2020domain}. 
StyleGAN \cite{karras2019style,karras2020analyzing,karras2021alias} achieves high visual quality and fidelity through its disentangled operations in the latent space $\mathcal{W}$. 
Corresponding researches \cite{yang2021semantic,collins2020editing,shen2020interpreting} have also demonstrated the disentanglement properties of the latent space $\mathcal{W}$. 
This property is particularly important for image manipulation with GAN latent spaces, thus making the study of pre-trained StyleGAN for downstream tasks a hot topic.

\begin{figure}[t]
\centering
\setlength{\abovecaptionskip}{1mm} %调整caption与图的距离
\setlength{\belowcaptionskip}{-0.6cm}%调整caption与下文的距离
\includegraphics[width=1\linewidth]{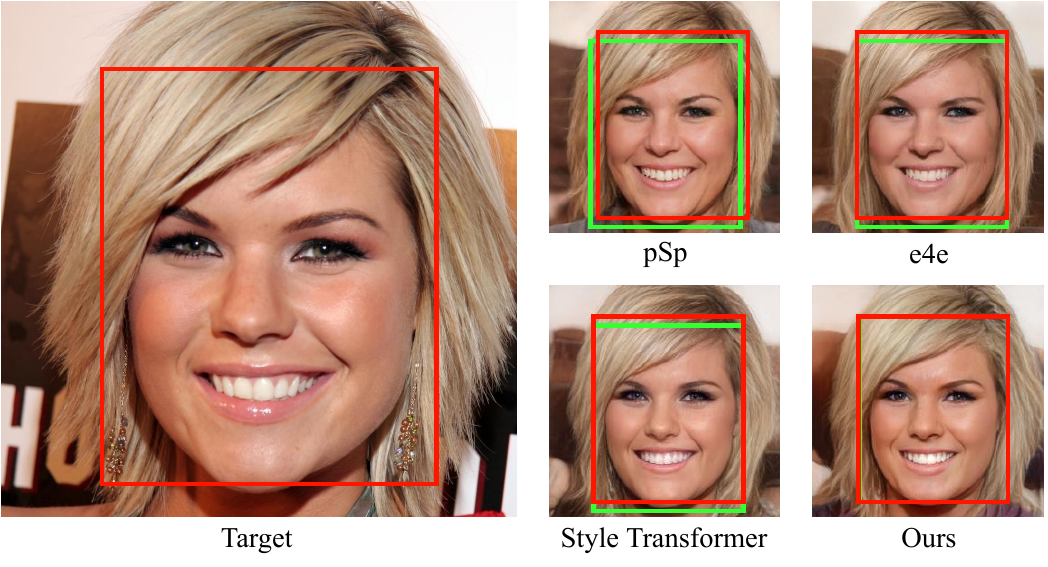}
\caption{Differences in image structure between convolutional backbone and Transformer backbone inversion results. The \textcolor{green}{green boxes} cover the facial outline of the inversion results for the different frameworks. The \textcolor{red}{red boxes} represents the size and location of the target's facial outline. The size and location of the facial outline of our results nearly overlap with the target.}
\label{fig0}
\end{figure}

In recent years, many works \cite{shen2020interpreting,jahanian2019steerability,tewari2020stylerig,harkonen2020ganspace} have verified the feasibility of performing image operations with latent space $\mathcal{W}$. 
However, studies \cite{abdal2019image2stylegan,richardson2021encoding} have shown that the latent space $\mathcal{W}$ representation is limited and that the images are inaccurate for mapping to $\mathcal{W}$ space. 
Inspired by Abdal et al. \cite{abdal2019image2stylegan}, a series of works \cite{abdal2019image2stylegan,abdal2020image2stylegan++,abdal2021styleflow,pbayliesstyleganencoder,zhu2020domain}, represented by pSp \cite{richardson2021encoding} and e4e \cite{tov2021designing}, 
invert StyleGAN in $\mathcal{W}+$ space and show encouraging results. 
In addition, Wang et al. \cite{wang2022high} and Roich et al. \cite{roich2022pivotal} further explore the inversion of image-specific details such as lighting, background, and makeup from different perspectives to achieve more realistic image editing. 
However, most of the current image inversion algorithms use the CNNs structure, which leads them to suffer from the inherent disadvantages of CNNs. Some work in recent years \cite{liu2023delving} indicates that the local receptive field of convolution leads to difficulties for the model to capture long-range dependencies and learn the global structure of the object. This affects the quality of the inversion image structure (e.g., the size of the facial outline and the location of the face in the image). Although Style Transformer \cite{hu2022style} uses the Transformer branch in its network design to achieve a more realistic inversion, we argue that there is still more application room for Transformer in the image inversion. As shown in Figure~\ref{fig0}, although the inversion framework with convolutional backbone inverts well in local details, the inversion results have obvious facial misalignment and facial outline differences compared with the target, while the inversion framework with Transformer can fit the global structure of the target properly.

\begin{figure*}
\centering
\setlength{\abovecaptionskip}{1mm} %调整caption与图的距离
\includegraphics[width=1\linewidth]{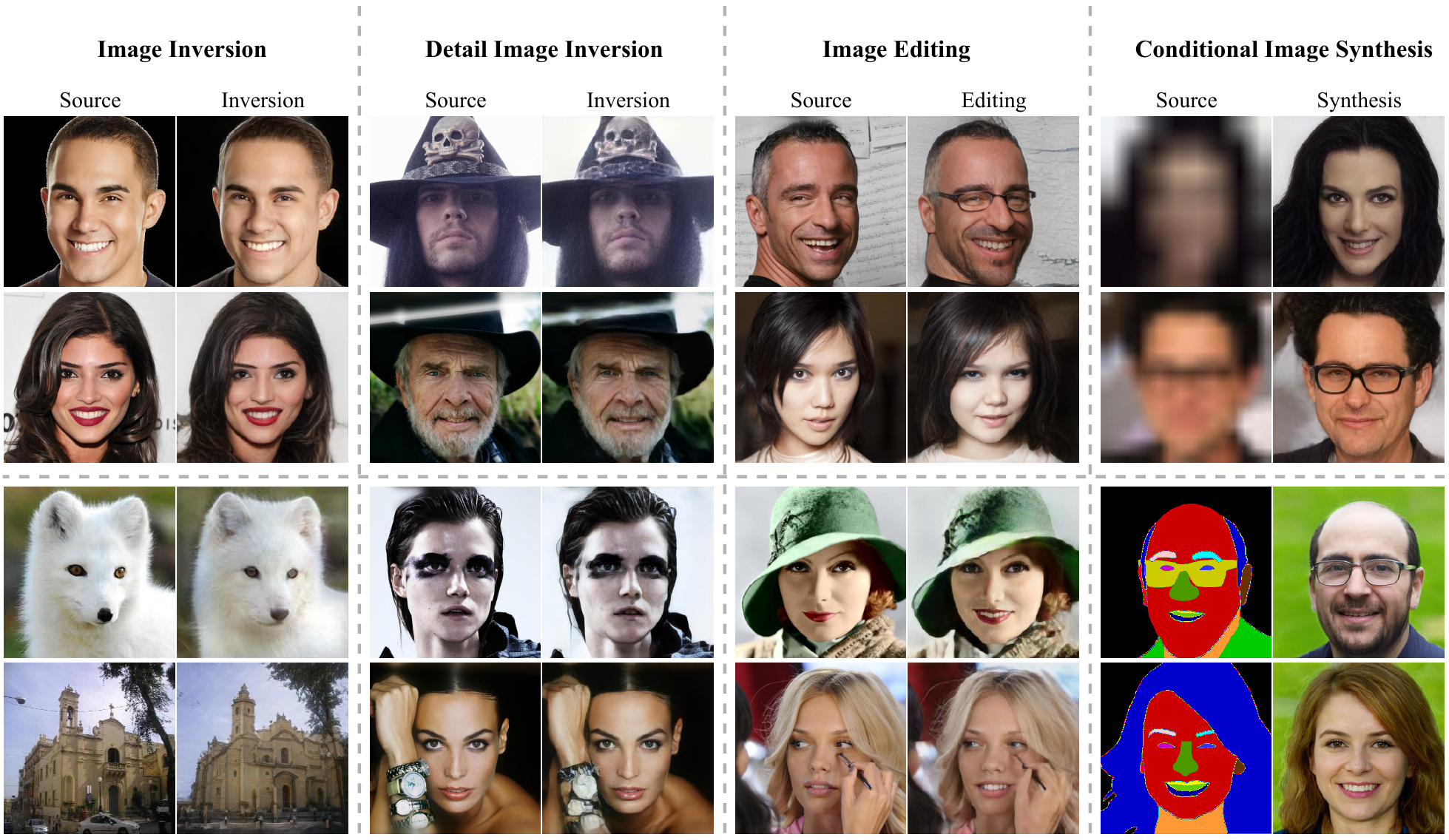}
\captionof{figure}{SwinStyleformer can perform well in image inversion and several tasks related to it. Examples include facial image inversion, image inversion on different domains, image inversion for specific details, image editing, image editing for specific details, image super resolution, and face from semantic segmentation map.}
\label{fig1}
\end{figure*}

\begin{figure}[t]
\centering
\setlength{\abovecaptionskip}{1mm} %调整caption与图的距离
\setlength{\belowcaptionskip}{-0.6cm}%调整caption与下文的距离
\includegraphics[width=1\linewidth]{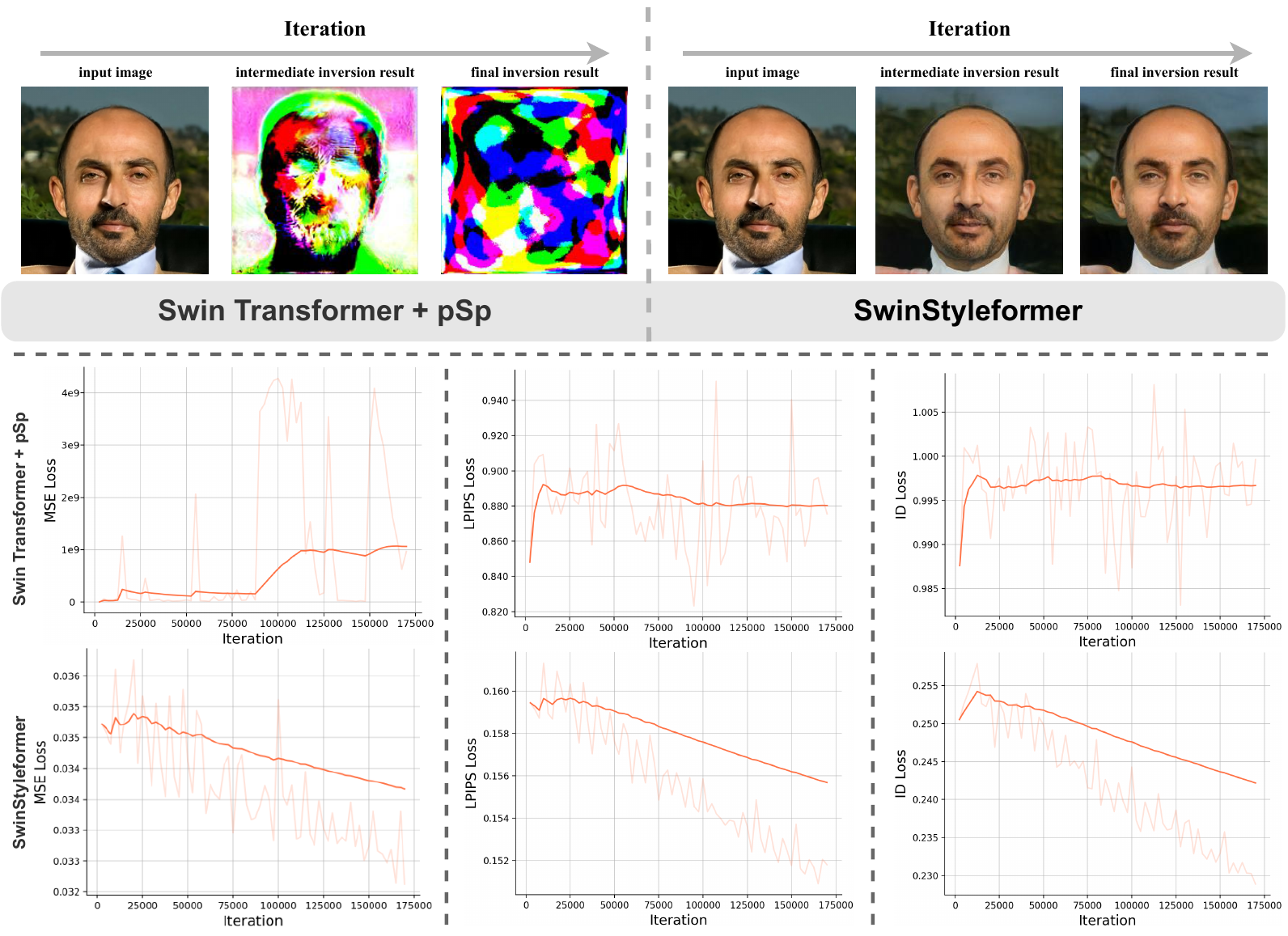}
\caption{Comparison of pSp with Swin Transformer backbone and SwinStyleformer.}
\label{fig2}
\end{figure}

   However, Transformer can't directly migrate to image inversion. 
When we replace the backbone of pSp with the generic Transformer backbone Swin Transformer \cite{liu2021swin}, we find the inversion fails (see Figure~\ref{fig2}). 
This phenomenon mainly stems from the differences between CNNs and Transformers which are specifically expressed in image inversion in the following three points. 
Firstly, the Transformer's self-attention weights tend to focus more on the image structure and ignore local details. 
Secondly, Transformer is not equipped with CNNs robust multi-scale design. 
Multi-scale modelling allows the model to capture information at different levels of granularity to deepen the model's overall understanding of the image, thus giving the model a balance between global structure and local detail. 
Thirdly, there is a distribution difference between the latent code extracted by Transformer and the StyleGAN style vector.

    To address these differences, we introduce SwinStyleformer in this paper, which is the first image inversion network with a pure Transformer structure. 
To model local detail, SwinStyleformer investigates the application of the smaller window size Swin Transformer to image inversion. 
More importantly, we design the Transformer module based on learnable queries as the basic unit for SwinStyleformer to extract latent codes in $\mathcal{W}+$ space. 
As learnable queries are not linear mappings of input features but random learnable vectors independent of the individual windows of the Swin Transformer, a Transformer module based on learnable queries is more adaptable and flexible than a self-attention Transformer module. 
This flexibility allows the model to update the attention weights according to the specific task to compensate for the Transformer's lack of recovery of image detail and to adapt to different image inversion situations.
Although the hierarchical structure of the Swin Transformer offers a multi-scale design, it can't adequately meet the Transformer's inversion requirements. 
Thus we design multi-scale connections for the standard feature pyramids on the Swin Transformer backbone. 
Feature maps at different scales have residual connections to feature maps at all previous scales. 
Meanwhile, we design the distribution alignment loss as well as the inversion discriminator specifically for image inversion. 
The distribution alignment loss and inversion discriminator reduce the distribution difference between the latent code and style vector from the latent distribution perspective and data distribution perspective respectively. 
Based on these designs, our SwinStyleformer successfully solves the Transformer's inversion failure issue and demonstrates SOTA performance and robustness in image inversion and several related vision tasks.
    
    The main contributions of this paper are: (1). We present SwinStyleformer, which is the first pure Transformer structured image inversion network.
It proves that Transformer is feasible in image inversion.
(2). SwinStyleformer successfully solves the Transformer's inversion failure issue.
(3). SwinStyleformer demonstrates SOTA performance and robustness in image inversion and several related vision tasks.

\section{Related Work}
    
\subsection{Image inversion}
    
    GANs \cite{goodfellow2014generative} show encouraging results on image generation tasks. Several studies \cite{goetschalckx2019ganalyze,jahanian2019steerability,shen2020interpreting,yang2021semantic} have found rich semantic information in the latent space $\mathcal{W}$ of GANs.
This has led to a series of image operations utilizing the latent space of GANs. 
The current popular research methods are mainly divided into encoder-based image inversion methods 
\cite{perarnau2016invertible,creswell2018inverting,pidhorskyi2020adversarial,nitzan2020disentangling,richardson2021encoding,tov2021designing,hu2022style,wang2022high,xu2022gh,guan2020collaborative,alaluf2021restyle,pidhorskyi2020adversarial} and direct optimization latent code image inversion methods \cite{lipton2017precise,creswell2018inverting,abdal2019image2stylegan,abdal2020image2stylegan++,huh2020transforming}.  

\begin{figure*}
        \centering
        \setlength{\abovecaptionskip}{1mm} %调整caption与图的距离
        \setlength{\belowcaptionskip}{-4mm}%调整caption与下文的距离
        \includegraphics[width=1\linewidth]{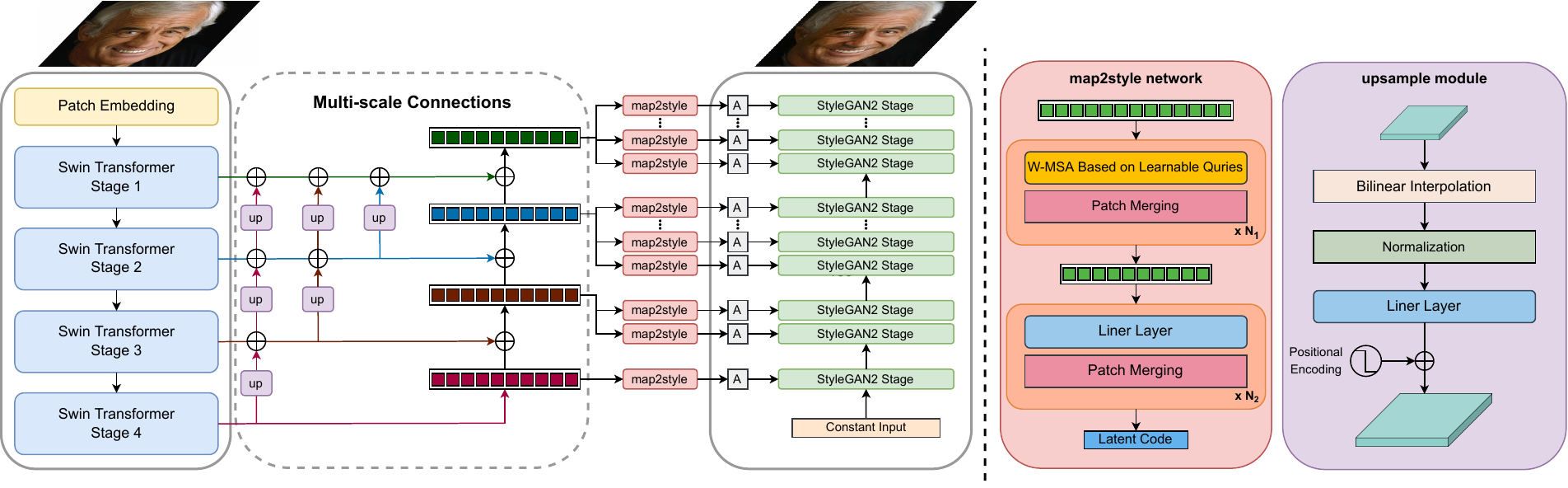}
        \caption{SwinStyleformer overall architecture. $A$ denotes the affine transformation corresponding to the latent code. $N_1$, $N_2$ denote the depth required for the sequence of tokens to the length of 16 and 1, respectively.}
        \label{fig3}
        \end{figure*}
    
\subsection{Latent Space Editing}
    
    Many works have investigated image editing with the latent space of GANs. 
\cite{harkonen2020ganspace,shen2021closed,voynov2020unsupervised,yuksel2021latentclr} search for the main direction controlling GANs synthesis in the latent space $\mathcal{W}$ with an unsupervised approach. 
Among them, GANSpace \cite{harkonen2020ganspace} and Sefa \cite{shen2021closed} mainly adopt the PCA \cite{shlens2014tutorial}, and LatentLCR \cite{yuksel2021latentclr} employs contrastive learning. 
Meanwhile, supervised learning based on attribute labels also plays an important role in the image editing of GANs. 
InterfaceGAN \cite{shen2020interpreting} utilizes a binary SVM \cite{cortes1995support} to separate bounding hyperplanes of attribute labels. 
Wang et al. \cite{wang2021attribute} suggested controlling image editing by $\mathcal{S}$ space composed of affine layers. 
StyleFlow \cite{abdal2021styleflow} successfully edits images based on conditions by utilizing continuous normalizing flows.
    
    \section{SwinStyleformer}

    \subsection{SwinStyleformer Framework}
    
    As shown in Figure~\ref{fig3}, to achieve efficient image inversion on $\mathcal{W}+$ space, SwinStyleformer follows the pipeline of pSp \cite{richardson2021encoding}. 
We redesigned the internal structure of pSp in order to compensate for the shortcomings of CNNs backbone to achieve higher quality inversion and to address the inversion failures caused by the difference between the Transformer and CNNs. 
We employ the generic Transformer framework Swin Transformer \cite{liu2021swin} as the backbone of SwinStyleformer to extract image features at the coarse, medium, fine, and finest levels. 
For better local modeling and to achieve a balance of efficiency, Swin Transformer adopts a window size of 2 for the first two stages and keeps the window size at 8 for the last two stages. 
Figure~\ref{appendixfig1} provides the basis. On the feature pyramid with Swin Transformer, feature maps of different granularity are aggregated in multi-scale connections for latent code extraction. 
For the four levels of feature maps output from the pyramid network, SwinStyleformer extracts latent codes using an intermediate network map2style which utilizes the Transformer module based on learnable queries as the basic unit. Finally, the latent codes aligned with the hierarchical representation are fed into the generator according to their scale to obtain the output image. The study in Table~\ref{tab17} also compares the influence of the convolutional backbone and advanced convolutional design with the Transformer backbone on the image inversion task, further demonstrating the contribution of the transformer structure to image inversion.
    
\subsection{SwinStyleformer Core Designs}
    
    \begin{figure}[h]
    \centering
    \setlength{\abovecaptionskip}{1mm} %调整caption与图的距离
    \setlength{\belowcaptionskip}{-0.1cm}%调整caption与下文的距离
    \includegraphics[width=1\linewidth]{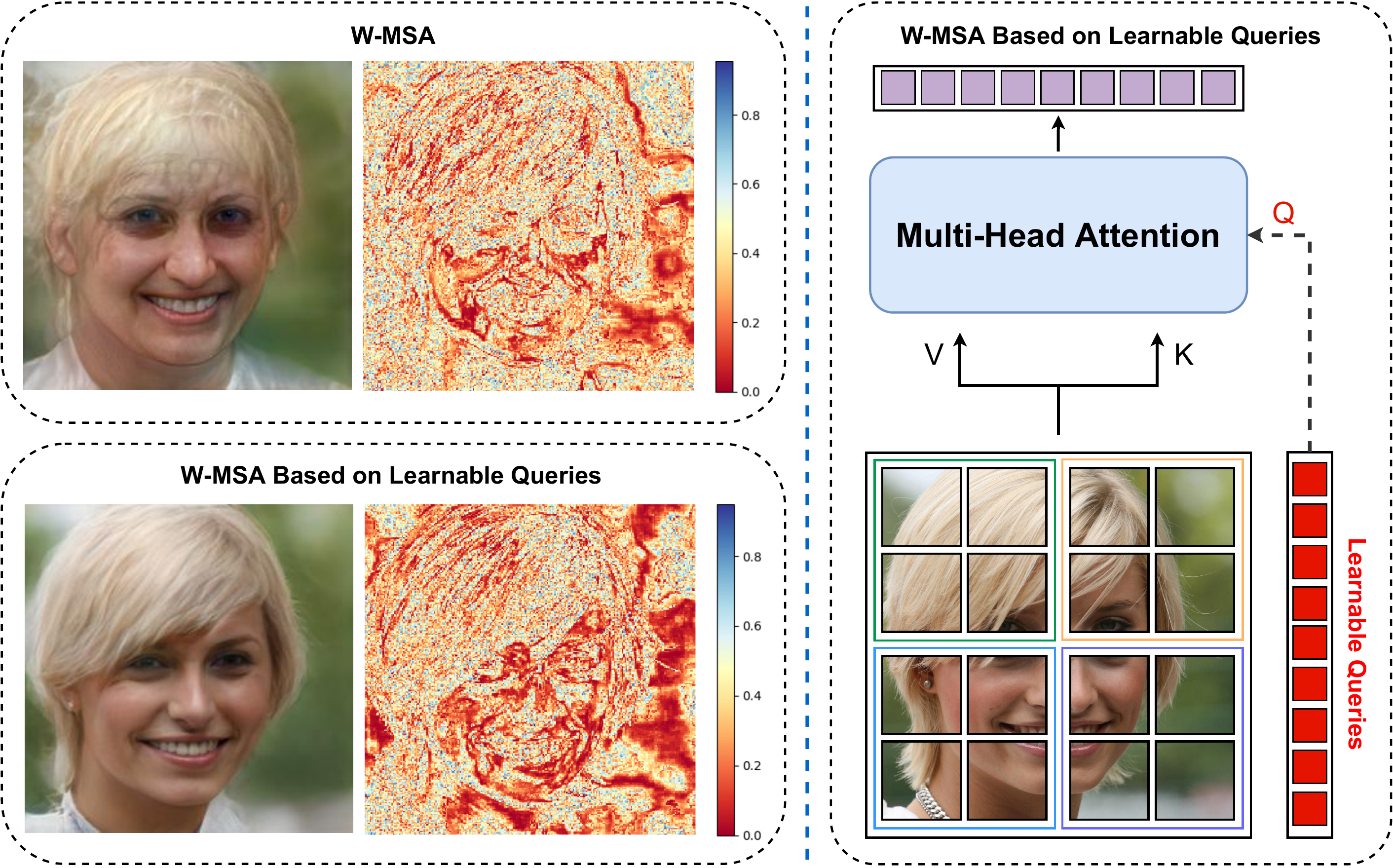}
    \caption{Overview of the W-MSA based on learnable queries, heat map of W-MSA and W-MSA based on learnable queries. 
We visualize the heat map with the difference between the inversion image and the input image to show the focused inversion region. 
It can be found that our method increases the attention to image details while retaining the attention to image structure.}
    \label{fig4}
    \end{figure}

    \paragraph{Multi-scale Connections} In terms of multi-scale, despite Swin Transformer's hierarchical design, it still falls short of convolution to some degree, which naturally has robust multi-scale properties. 
Multi-scale designs play an important role in image inversion. 
By introducing a multi-scale design, SwinStyleformer can ensure that important visual information is captured regardless of the size of the object. 
Meanwhile, the multi-scale design allows the SwinStyleformer to capture both local detail and global context to gain holistic learning of the image. 
The above two factors motivate SwinStyleformer to preserve detail while inverting the structure. 
Thus, we design multi-scale connections in SwinStyleformer's feature pyramid to further introduce multi-scale properties. 
Feature maps at different scales are residually connected to feature maps at all previous scales in standard feature pyramids (see Figure~\ref{fig3}). Table~\ref{appendixtab2} shows that multi-scale connections are indispensable.
    
    \paragraph{Upsample Module} SwinStyleformer also designs upsampling modules for different level feature maps in multi-scale connections. 
The upsampling module still performs the bilinear interpolation of the input feature map in the same way as pSp. 
Yet, pSp requires convolution to align the feature map channels. 
Motivated by Patch Expanding in Swin-Unet \cite{cao2022swin}, SwinStyleformer applies linear layers to align feature map channels and adds normalization operations. 
Studies \cite{anokhin2021image,lee2021vitgan} have demonstrated that for generation tasks, letting the model be aware of the absolute position is crucial for the synthesis of specific components ({\em e.g.} the eye). 
Thus, SwinStyleformer adds absolute position encoding to the upsampled feature maps and controls the need for spatial coordinates itself with learnable parameters. 
The ablation studies in Table~\ref{appendixtab4} provide support for our above design.
    
\paragraph{Multi-head Attention Based on Learnable Queries} SwinStyleformer first considers Swin Transformer's window attention module as the basic unit for the intermediate network map2style. However, the stacking of window attention modules severely affects SwinStyleformer's inversion efficiency and causes significant artifacts. The heat map in Figure~\ref{fig4} shows that the artifact mainly arises from the window multi-head self-attention (W-MSA) mechanism based on queries after linear mapping of input features over-set attention to the structure. To solve the above issues, we design the Transformer module based on learnable queries for the map2style network. The core of this module is W-MSA based on learnable queries. For the attention mechanism, its queries, keys, and values are all based on projections from the same input. Thus, in the inversion task, this fixed-query attention is more likely to focus on the basic structure of the image while ignoring subtle features during global modeling. Compared to W-MSA, W-MSA based on learnable queries are randomly initialized and updated based on losses. So, learnable queries prompt the model to focus on the difference between the result and the target, thus boosting the areas of poorer inversion, which lead the model will not be limited to focus on the structure, which enables our method to become more adaptive and flexible. It allows the model to learn and update the attention weights according to the specific task and allows the model to selectively focus on relevant features in the input image as required. This flexibility and adaptability are not available with a fixed query. Ultimately, with relatively less attention paid to structure, learnable queries can allocate more attention to important inversion elements (e.g., local details), thus resolving artifacts and becoming a key component of SwinStyleformer's solution to inversion failure.  
Meanwhile, the efficiency issue is addressed as each window shares learnable query weights. 
See Table~\ref{appendixtab1} for more discussion.
    
    \paragraph{Map2style Network} Apart from the Transformer module based on learnable queries, SwinStyleformer offers other designs for map2style network. 
Firstly, we follow the Swin Transformer's Patch Merging operation to downsample the embedded token sequence. 
Secondly, given the efficiency of inversion, for embedded token sequence after downsampling to a length of 16, we no longer let them self-focus based on learnable queries. 
We can process them well with the linear layer. The above designs contribute to SwinStyleformer's success in encoding latent code in $\mathcal{W}+$ space. 
This is supported by the results of the ablation experiments and visualization in Figure~\ref{appendixfig2}, respectively.
    
    \subsection{SwinStyleformer Loss Functions}
    
    \begin{figure}[t]
    \centering
    \setlength{\abovecaptionskip}{1mm} %调整caption与图的距离
    \setlength{\belowcaptionskip}{-0.4cm}%调整caption与下文的距离
    \includegraphics[width=1\linewidth]{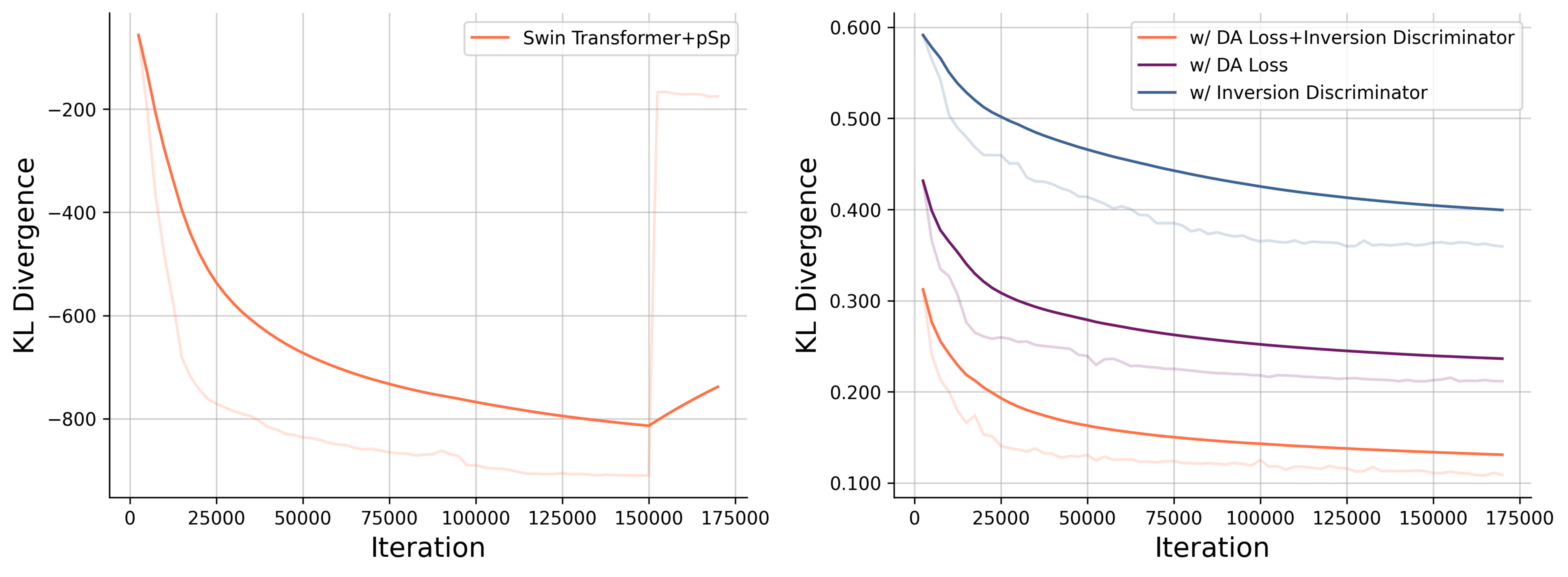}
    \caption{Distribution differences between latent codes and style vectors under different baselines.}
    \label{fig5}
    \end{figure}

    \begin{figure*}[t]
    \centering
    \setlength{\abovecaptionskip}{1mm} %调整caption与图的距离
    \setlength{\belowcaptionskip}{0cm}%调整caption与下文的距离
    \includegraphics[width=1\linewidth]{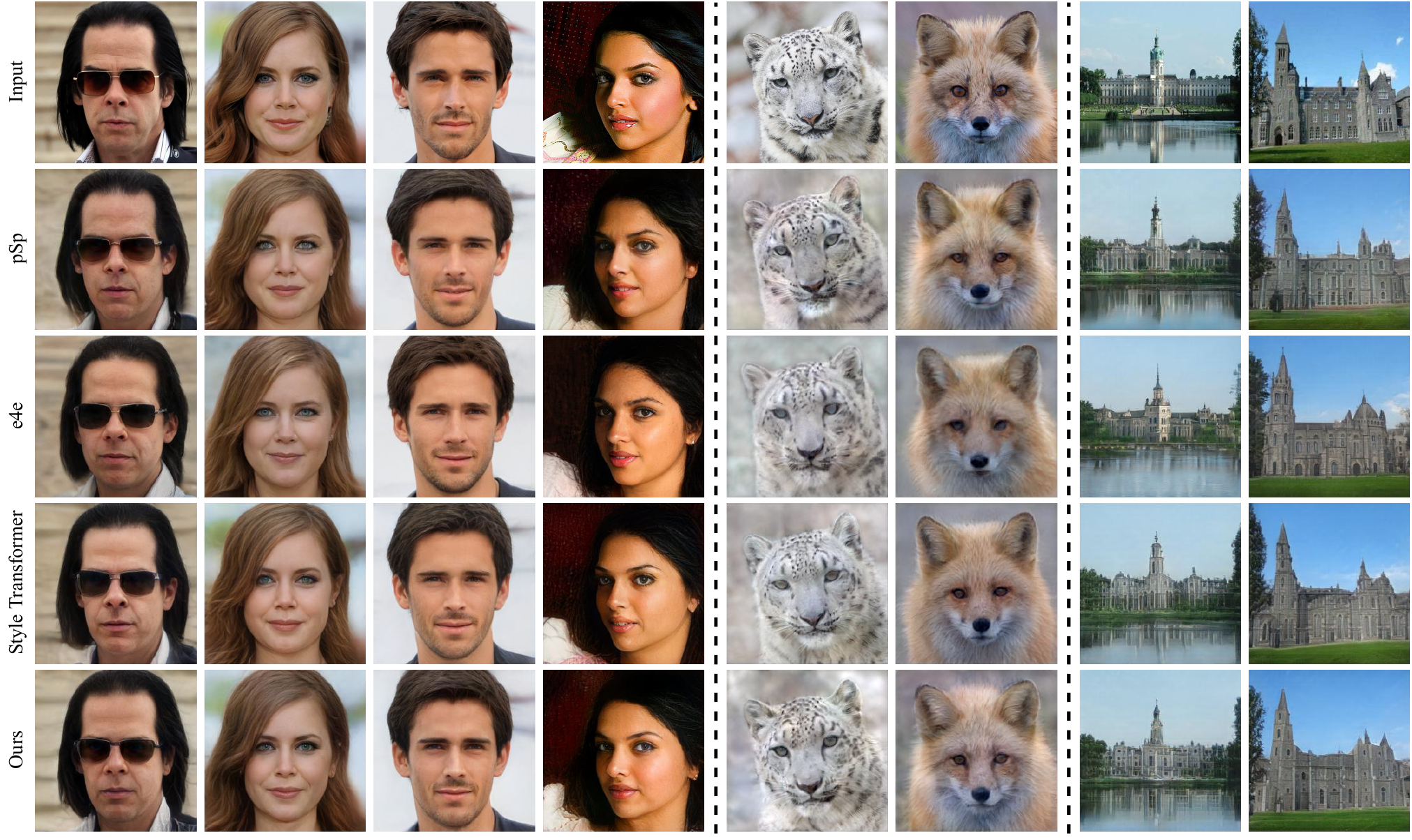}
    \caption{Inversion results of SwinStyleformer on different domains.}
    \label{fig7}
    \end{figure*}
    
    \paragraph{Distribution Alignment Loss} Figure~\ref{fig5} indicates that there are significant distribution differences between the latent code obtained by Swin Transformer and the StyleGAN style vector. Pidhorskyi et al. 
\cite{pidhorskyi2020adversarial} showed that in an encoder-based inversion framework, the encoder needs to map the input data onto a space represented by the latent distribution, while the generator needs to map the latent code onto a space represented by the data distribution. 
Thus the above issue can be addressed in both data distribution and latent distribution. 
We first investigate the solution from the latent distribution. 
SwinStyleformer first uses the non-linear mapping network in StyleGAN to map the latent code $z$ in the latent space $Z$ to style vector $w \in W$. 
We then employ the softmax function to obtain the distribution of latent code and the distribution of style vector. 
For the above distributions, SwinStyleformer directly constrains them with KL \cite{hershey2007approximating} scatter. 
The distribution alignment loss based on KL scatter prompts SwinStyleformer to better adapt the latent code to closely match the distribution of the style vector for the generator to gain realistic results.
    
    % \begin{figure}[t]
    % \centering
    % \setlength{\abovecaptionskip}{1mm} %调整caption与图的距离
    % \setlength{\belowcaptionskip}{-0.4cm}%调整caption与下文的距离
    % \includegraphics[width=1\linewidth]{./fig/fig6.pdf}
    % \caption{Inversion discriminator framework.}
    % \label{fig6}
    % \end{figure}

    \begin{figure*}[t]
        \centering
        \setlength{\abovecaptionskip}{1mm} %调整caption与图的距离
        \setlength{\belowcaptionskip}{-0cm}%调整caption与下文的距离
        \includegraphics[width=1\linewidth]{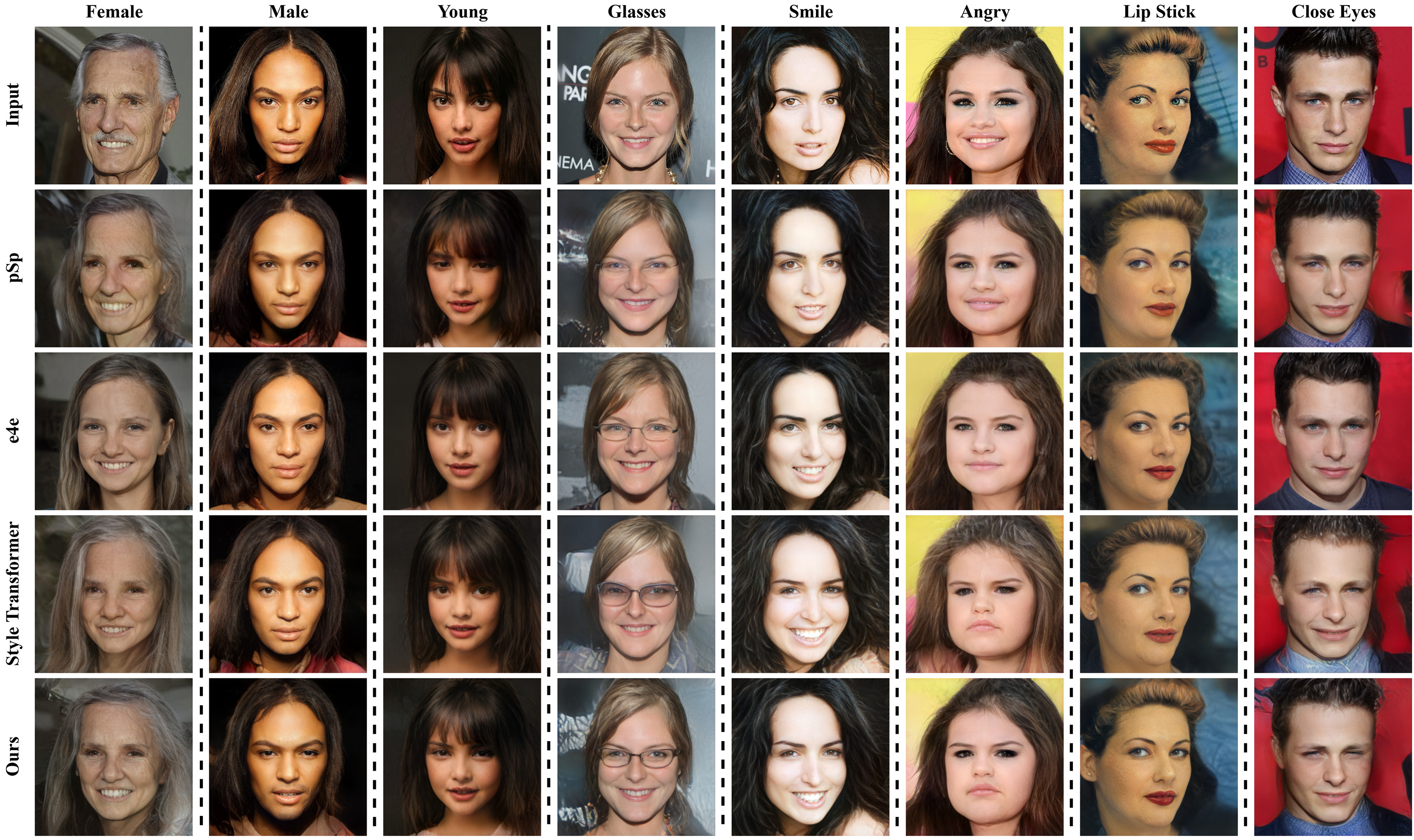}
        \caption{Qualitative comparison results of image editing. We show the results of image editing with several editing directions.}
        \label{fig8}
        \end{figure*}

    \begin{table*}[h]
    \centering
    \scriptsize
    \setlength{\abovecaptionskip}{0.1cm} %调整caption与图的距离
    \setlength{\belowcaptionskip}{-0.4cm}%调整caption与下文的距离
    % \resizebox{17.2cm}{!}{
    \setlength\tabcolsep{1.4mm}
    \begin{tabular}{c||ccccc|ccccc|ccccc}
    %\toprule[1.4pt]
    \hline\thickhline
    \rowcolor{mygray}
         & \multicolumn{5}{c|}{Facial Domain}                                                                          & \multicolumn{5}{c|}{Animal Domain}                                                                        & \multicolumn{5}{c}{Church Domain} \\
         \cline{2-16}
         \rowcolor{mygray}
         \multirow{-2}[0]{*}{Method}& MSE $\downarrow$  & LPIPS $\downarrow$  & FID $\downarrow$  & PSNR $\uparrow$  & SSIM $\uparrow$  & MSE $\downarrow$ & LPIPS $\downarrow$ & FID $\downarrow$ & PSNR $\uparrow$  & SSIM $\uparrow$  & MSE $\downarrow$  & LPIPS $\downarrow$ & FID $\downarrow$ & PSNR $\uparrow$   & SSIM $\uparrow$ \\
         \hline\hline
         ALAE  & 0.182             & 0.430             & 24.860             & 10.657            &
         0.398             & 0.379             & 0.672             & 51.109            & 9.033             &      0.287             & 0.357             & 0.631             & 49.465            & 11.178            &
         0.309 \\
         pSp & 0.038             & 0.176             & 37.561            & 20.948            & 0.603             & 0.085             & 0.280             & 13.732            & 17.031            & 0.281             & 0.098             & 0.325             & 56.061            & 16.093            & 0.327 \\
        e4e & 0.050              & 0.207             & 43.295            & 19.138            & 0.518             & 0.090              & 0.325             & 14.670             & 16.238            & 0.247             & 0.129             & 0.394             & 58.503            & 14.807            & 0.265 \\
        Style Transformer & 0.035             & 0.169             & 36.362            & 20.918            & 0.601             & 0.082             & 0.278             & 13.416            & 17.139            & 0.281             & 0.096             & 0.317             & 62.926            & 16.126            & 0.344 \\
        CLCAE & 0.034 & 0.159 &  32.530 & 21.232 & 0.596 & 0.083 & 0.277 & 12.851 & 17.336 & 0.286 & 0.096 & 0.313 & 57.130 & 16.049 & 0.340\\
        \hline
        Ours              & \textbf{0.032}    & \textbf{0.159}    & \textbf{34.218}   & \textbf{21.274}   & \textbf{0.617}    & \textbf{0.080}     & \textbf{0.275}    & \textbf{12.459}   & \textbf{17.215}   & \textbf{0.283}    & \textbf{0.094}    & \textbf{0.309}    & \textbf{55.592}   & \textbf{16.381}   & \textbf{0.352} \\

    \hline\thickhline
    \end{tabular}%}
    \caption{\footnotesize{Quantitative comparison results of image inversion for different domains.}}
    \label{tab1}
    \end{table*}

    \paragraph{Inversion Discriminator } Many efforts \cite{donahue2016adversarial,dumoulin2016adversarially,srivastava2017veegan,ulyanov2018takes,huang2018introvae} design adversarial games to ensure the generator output matches the training data distribution. 
The latent code can be motivated to approximate the style vector distribution by scaling down the data distribution between the inversion image and the input image. 
However, the existing discriminators are better suited to image generation. 
For image inversion, the adversarial game requires not only the generation of a realistic image but also the ability to restore the input image. 
Thus it is necessary to design a more demanding discriminator to reduce the distribution difference between the inversion image and the input image in image inversion. 
Motivated by contrastive learning work \cite{he2020momentum,chen2020improved,chen2021mocov3}, SwinStyleformer designs an inversion discriminator specifically for image inversion tasks. 
The inversion discriminator employs StyleGAN's discriminator framework as the backbone to encode the inversion images. 
The inversion discriminator also has a momentum encoder that encodes the input image to amplify the differences between the encodings. 
    We have the following objectives:
    \begin{equation}
        \begin{split}
        &\min_{D} L(D) = \mathbb{E}_{x\sim P_x}\left((D(x,x) - 1)^2 \right. \\
        &\left. + (D(G(E(x)),x) + 1)^2\right), \\
        &\min_{E} L(E) = \mathbb{E}_{x\sim P_x}\left(D\left(G(E(x),x\right) - 1\right)^2), \\
        &D(x,x) = \cos(m_q(x), m_k(x)), \\
        &D(G(E(x)),x) = \cos(m_q(G(E(x))), m_k(x)),
        \end{split}
        \label{eq1}
    \end{equation}
where $D$ and $E$ denote the inversion discriminator and SwinStyleformer respectively. 
$G$ is StyleGAN, and $x$ is the input.   
$cos(.)$ denotes cosine similarity.  
$m_q$ and $m_k$ denote the source encoder and the momentum encoder respectively. 
The momentum update weight is 0.999. 
The inversion discriminator can effectively meet the requirements of inversion while reducing the distribution differences. The total loss is described in the Appendix.
Section~\ref{ablation} provides relevant discussions.
    
    %\paragraph{Total Loss. }In addition to the distribution alignment loss and the inversion discriminator. We follow the losses of pSp to constrain inversion training, including $L2$ loss, LPIPS \cite{zhang2018unreasonable} loss for perceptual similarity, and ID \cite{richardson2021encoding} loss for preserving input identity.
%\begin{equation}
    %\begin{split}
%&L({\rm{x}}) = {\lambda _1}{L_2}({\rm{x}}) + {\lambda _2}{L_{{\rm{LPIPS}}}}({\rm{x}}) \\
%&+ {\lambda _3}{L_{{\rm{ID}}}}({\rm{x}}) + {\lambda _4}{L_{{\rm{DA}}}}({\rm{x}}) + {\lambda _5}{L_{{\rm{adv}}}}({\rm{x}}),
%\end{split}
%\label{eq2}
%\end{equation}
%where $L_{DA}(.)$ is the distribution alignment loss and $L_{adv}(.)$ is the adversarial loss. 
%$\lambda_1$, $\lambda_2$, $\lambda_3$, $\lambda_4$, $\lambda_5$ are the hyperparameters. 
%These loss functions help SwinStyleformer to achieve effective inversion while reducing the distribution difference between the latent code and the style vector. 
%Appendix \textcolor{red}{B} describes the settings of the individual hyperparameters.
    
\section{Experiments}

To explore the effectiveness of SwinStyleformer, we evaluate it in image inversion and other tasks. Specific details and more results can be found in the \textbf{Appendix}.

\subsection{Image Inversion}

We first test the inversion effectiveness of SwinStyleformer. 
In the image inversion, we compare SwinStyleformer with the current SOTA inversion algorithms \cite{pidhorskyi2020adversarial,richardson2021encoding,tov2021designing,hu2022style,liu2023delving}. To verify the robustness of SwinStyleformer, we test it in the facial domain, the animal domain, and the church domain respectively.

\paragraph{Quantitative results} Table~\ref{tab1} demonstrates the results of the SwinStyleformer quantitative evaluation. 
The images inverted by SwinStyleformer exhibit better perceptual similarity than others. 
We also measure the similarity with the FID \cite{heusel2017gans} to be independent of our loss function. 
A similar situation is observed with regard to the FID. SwinStyleformer gains better PSNR and SSIM. Quantitative indicators of the inversion for three different domains validate the robustness of SwinStyleformer.

\paragraph{Qualitative results} We visualize the performance of SwinStyleformer in different domain inversion tasks in Figure~\ref{fig7}. 
SwinStyleformer and the current SOTA image inversion algorithms perform well and are robust enough in several domains. 
Compared to them, SwinStyleformer performs better in details and structure {\em e.g.} hair color, eye shadow, and facial smile details in the second column of images or viewpoint and rosy cheeks in the fourth column of images. 
Moreover, SwinStyleformer shows richer detail in other fields.

\subsection{Image Editing}

We test the effectiveness of SwinStyleformer for image editing. For the above editing tasks, we use the editing directions from InterFaceGAN \cite{shen2020interpreting}. Our five workers perform human evaluations for face from segmentation map, image editing, and image editing for specific details, each of them reviews approximately 3K pairs for each task. For each operation, each method receives over 15K human judgments.

\begin{table}[h]
\centering
\scriptsize
\setlength{\abovecaptionskip}{1mm}
\setlength{\belowcaptionskip}{-4mm}
\setlength\tabcolsep{4mm}
\begin{tabular}[t]{c|ccc}
  \hline\thickhline
  \rowcolor{mygray}
  Method & FID $\downarrow$ & ID $\uparrow$ & Rates $\uparrow$\\
  \hline
  pSp \cite{richardson2021encoding} & 48.792 & 0.529 & 91.50\% \\
  e4e  \cite{tov2021designing} & 54.136 & 0.468 & 94.20\% \\
  Style Transformer \cite{hu2022style} & 47.105 & 0.510 & 91.70\% \\
  \hline
  Ours & \textbf{43.089} & \textbf{0.573} & \textbf{95.10\%} \\
  \hline\thickhline
\end{tabular}
\caption{Quantitative comparison of image editing. ID means identity similarity using a
pre-trained facial recognition network~\cite{huang2020curricularface}. Rates indicate the results of the human evaluation.}
\label{tab13}
\end{table}

\begin{figure}[t]
\centering
\setlength{\abovecaptionskip}{1mm} %调整caption与图的距离
\setlength{\belowcaptionskip}{-0.4cm}%调整caption与下文的距离
\includegraphics[width=1\linewidth]{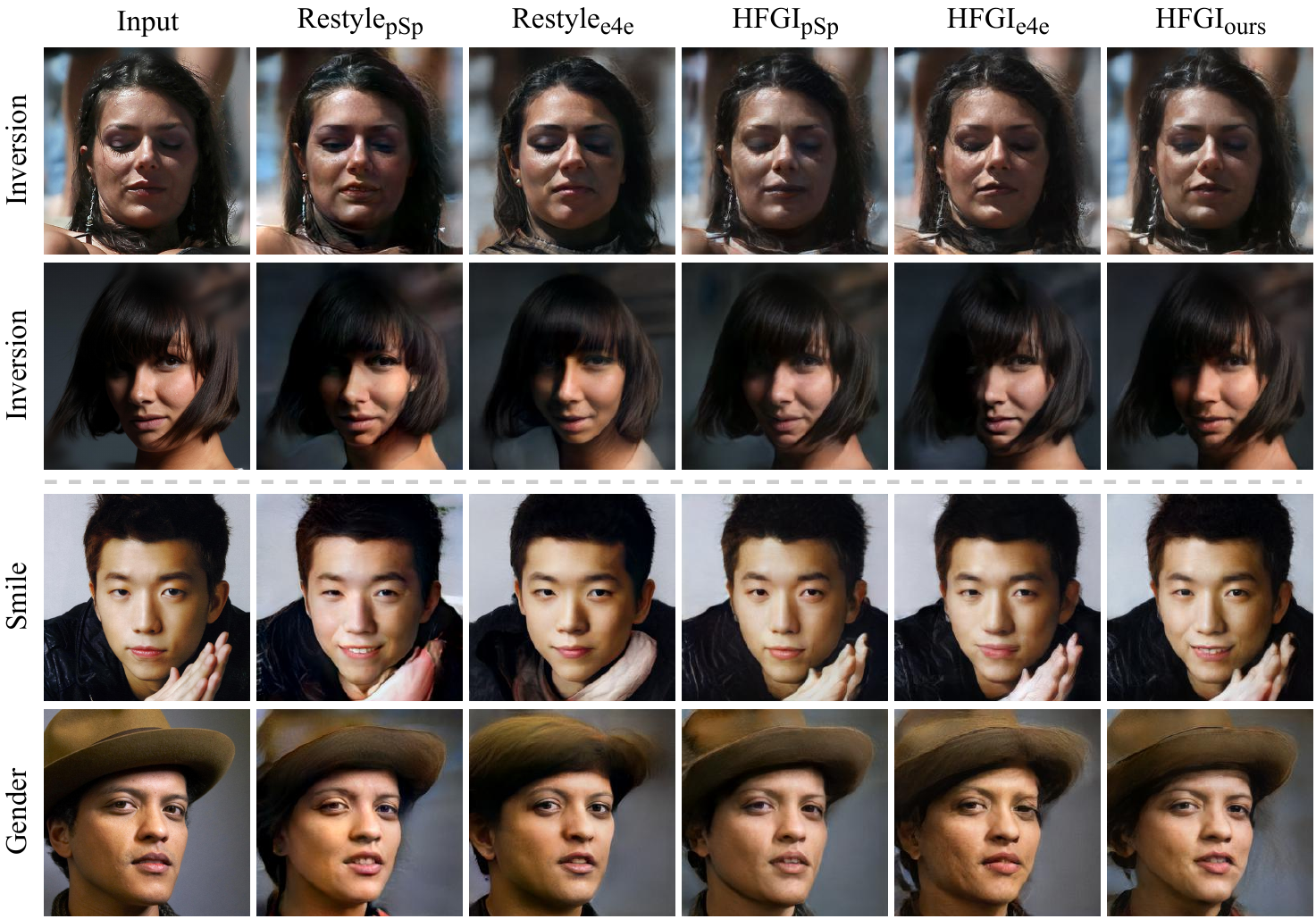}
\caption{Visual comparison of image inversion and image editing for specific details.}
\label{fig9}
\end{figure}

\begin{figure}[h]
\centering
\setlength{\abovecaptionskip}{1mm} %调整caption与图的距离
\setlength{\belowcaptionskip}{-0.5cm}%调整caption与下文的距离
\includegraphics[width=1\linewidth]{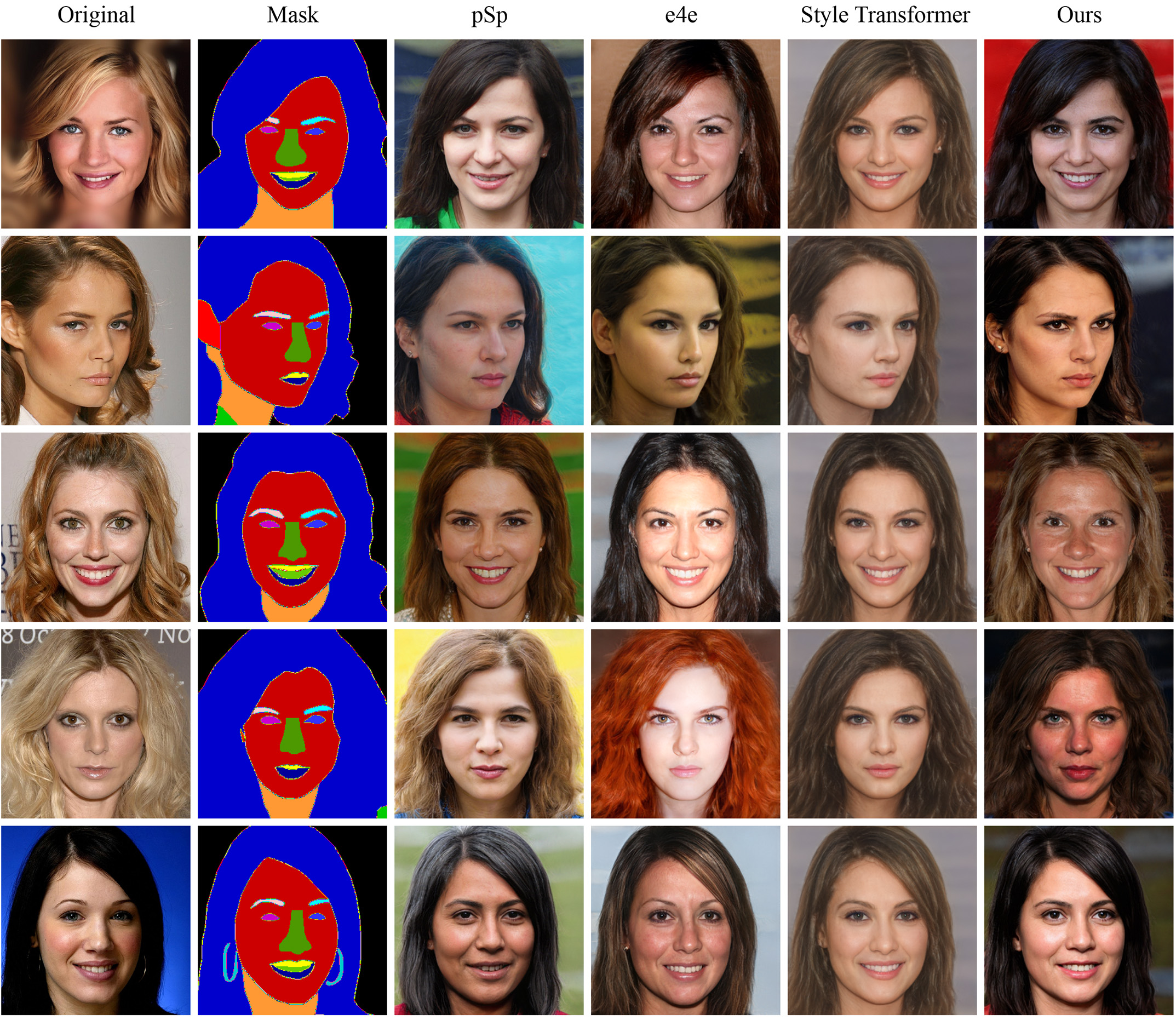}
\caption{Comparison of SwinStyleformer with several algorithms for semantic segmentation map to face.}
\label{fig10}
\end{figure}

\paragraph{Results} Figure~\ref{fig8} shows SwinStyleformer perform better on image editing task. 
For example, in the case of gender editing, the effect of our method is more obvious. On the edits of angry and closed eyes, both pSp and e4e edits are less apparent. 
Although the StyleTransformer effect is obvious, it makes a significant change compared to the input image. 
In comparison, our method achieves significant editing while keeping as similar as possible to the original image. Meanwhile, Table~\ref{tab13} provides a quantitative evaluation of image editing on CelebA-HQ~\cite{karras2017progressive}.

\subsection{Inversion and Editing for Specific Details}

Since low-rate latent codes in $\mathcal{W}+$ space are not sufficient to invert specific details ({\em e.g.}. make-up, lighting, and background), we investigate the contribution of SwinStyleformer to the inversion of specific details combined with the framework HFGI \cite{wang2022high}. 
We compare SwinStyleformer as the backbone of HFGI with other methods \cite{alaluf2021restyle,roich2022pivotal,alaluf2022hyperstyle} and HFGI of other backbones.

\begin{table}[h]
    \centering
    \scriptsize
    \setlength{\abovecaptionskip}{0.1cm} %调整caption与图的距离
    \setlength{\belowcaptionskip}{-0.3cm}%调整caption与下文的距离
    % \resizebox{17.2cm}{!}{
    \setlength\tabcolsep{2mm}
    \begin{tabular}{c|ccccc}
    %\toprule[1.4pt]
    \hline\thickhline
    \rowcolor{mygray}
    Method            & MSE $\downarrow$  & LPIPS $\downarrow$  & FID $\downarrow$  & PSNR $\uparrow$  & SSIM $\uparrow$ \\
    \hline
    PTI & 0.015 & 0.099 & 24.168 & 24.715 & 0.804 \\
    HyperStyle & 0.019 & 0.097 & 23.972 & 24.831 & 0.793 \\
    \hline
    Restyle$_{pSp}$     & 0.027             & 0.130              & 32.125            & 22.554            & 0.675 \\
    Restyle$_{e4e}$ & 0.042             & 0.184             & 33.873            & 20.192            & 0.568 \\
    \hline
    HFGI$_{pSp}$ & 0.016             & 0.106             & 24.715            & 24.642            & 0.789 \\
    HFGI$_{e4e}$ & 0.020              & 0.108             & 24.760             & 23.736            & 0.761 \\
    \hline
    HFGI$_{ours}$          & \textbf{0.014}    & \textbf{0.093}    & \textbf{22.322}   & \textbf{25.236}   & \textbf{0.815} \\
    \hline\thickhline
\end{tabular}%
    \caption{\footnotesize{Quantitative results of image inversion for specific details.}}
    \label{tab2}
    \end{table}

\paragraph{Quantitative results} We observe encouraging results in Table~\ref{tab2}. 
Compared with the previous approaches, the SwinStyleformer with HFGI achieves a higher fidelity inversion. 
Our method achieves SOTA in all metrics for image inversion for specific details. 
Comparative results with other backbones of HFGI also prove that our success is no accident. 
This suggests that our method is better suited as the backbone of the inversion framework for specific details.

\begin{table}[h]
    \centering
    \scriptsize
    \setlength{\abovecaptionskip}{1mm}
    \setlength{\belowcaptionskip}{-2mm}
    \setlength\tabcolsep{3mm}
    \begin{tabular}[t]{c|ccc}
      \hline\thickhline
      \rowcolor{mygray}
      Method & FID $\downarrow$ & ID $\uparrow$ & Rates $\uparrow$\\
      \hline
      Restyle$_{pSp}$ \cite{alaluf2021restyle} & 44.627 & 0.571 & 60.80\% \\
      Restyle$_{e4e}$ \cite{alaluf2021restyle} & 45.801 & 0.552 & 65.00\% \\
      HFGI$_{pSp}$ \cite{wang2022high} & 35.620 & 0.589 & 95.80\% \\
      HFGI$_{e4e}$ \cite{wang2022high} & 34.149 & 0.583 & 95.60\% \\
      \hline
      HFGI$_{ours}$ & \textbf{33.296} & \textbf{0.616} & \textbf{96.90\%} \\
      \hline\thickhline
    \end{tabular}%
   
    \caption{Quantitative results of image editing for specific details.}
    \label{tab14}
\end{table}

\paragraph{Qualitative results} Figure~\ref{fig9} shows the visualization results. 
Compared with the HFGI inversion results using other backbones, the HFGI combined with our method is closer to the input image in illumination and more obvious editing for specific details. 
Although Restyle inverts the lighting of the original image and makes the editing apparent, there are obvious artifacts in some details, such as hair and hand.

\subsection{Face from Segmentation Map}

Here, we evaluate SwinStyleformer to synthesize a face image from a segmentation map. 

\begin{table}[h]
    \centering
    \scriptsize
    \setlength{\abovecaptionskip}{1mm} %调整caption与图的距离
    \setlength{\belowcaptionskip}{-2mm}%调整caption与下文的距离
    \setlength\tabcolsep{2mm}
      \begin{tabular}{c|cccc}
        \hline\thickhline
        \rowcolor{mygray}
      Method            & pSp               & e4e               & Style Transformer  & Ours \\
      \hline
      Face from Segmentation      & 94.70\%           & 94.90\%           & 90.20\%           & \textbf{96.30\%} \\
      \hline\thickhline
    \end{tabular}%
   
    \caption{Human evaluation of Face from Segmentation.}
    \label{tab12}%
%yletransformer
  \end{table}%

\paragraph{Results. }Figure~\ref{fig10} shows the results of our method compared to pSp, e4e, and Style Transformer.  
Our method exhibits richer facial details. 
This appears more realistic for our results. 
Although baselines all successfully invert the facial images according to the mask requirements, they still fall short of the masked areas of the input images such as the nose in the second row and the eyes in the fourth row. 
Compared to these algorithms, SwinStyleformer is more accurate. % for the inversion of the mask region.

\subsection{Super Resolution}

SwinStyleformer can also perform image resolution tasks. We investigate the performance of SwinStyleformer with other algorithms at different downsampling ratios.

\begin{figure}[t]
\centering
\setlength{\abovecaptionskip}{1mm} %调整caption与图的距离
\setlength{\belowcaptionskip}{-0.4cm}%调整caption与下文的距离
\includegraphics[width=1\linewidth]{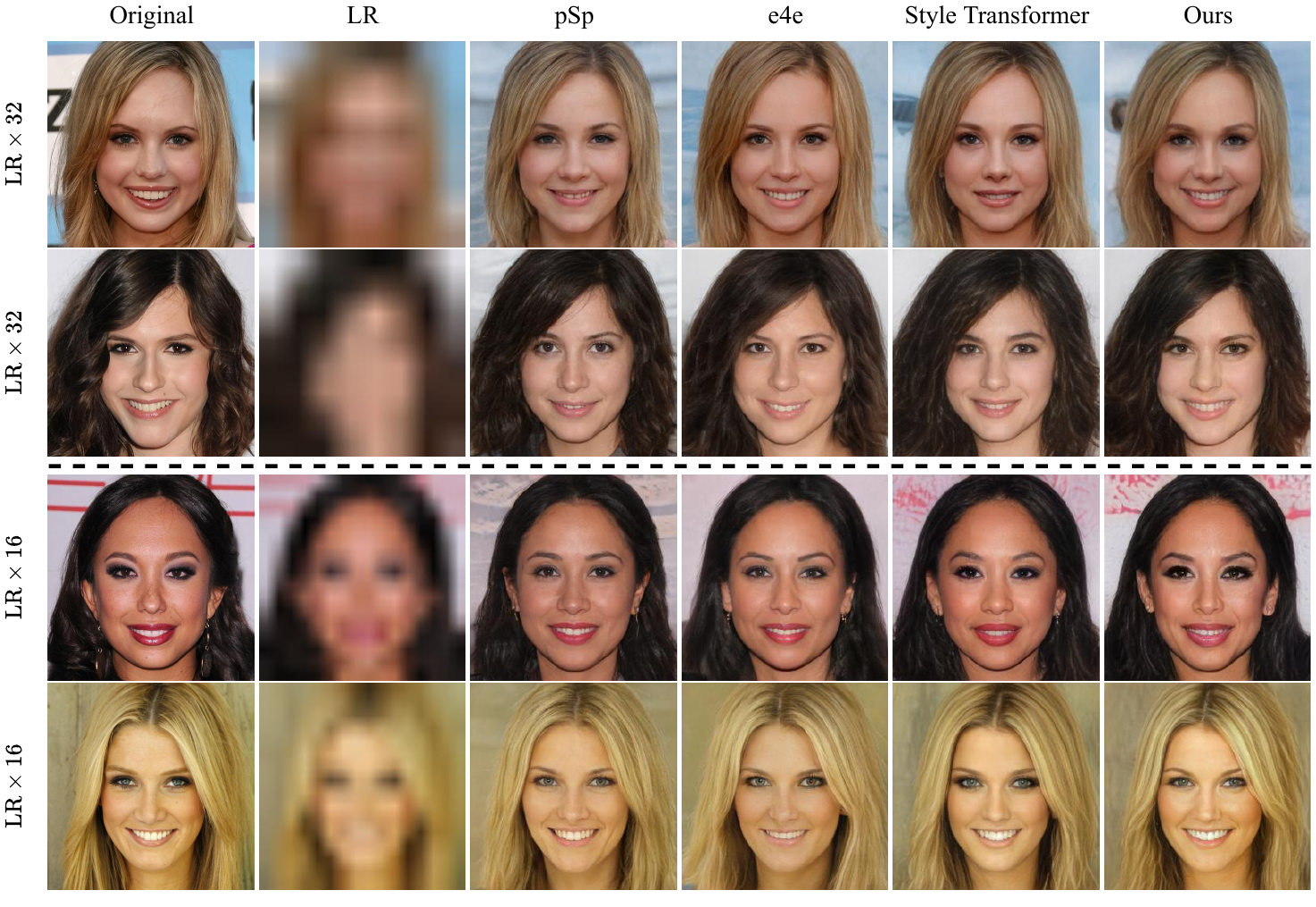}
\caption{Qualitative comparison of SwinStyleformer with baseline at $16 \times $ downsampling ratio and $32 \times $ downsampling ratio.}
\label{fig11}
\end{figure}

\begin{table}[h]
    \centering
    \scriptsize
    \setlength{\abovecaptionskip}{1mm} %调整caption与图的距离
    \setlength{\belowcaptionskip}{-2mm}%调整caption与下文的距离
    \setlength\tabcolsep{2mm}
    \resizebox{8.5cm}{!}{
      \begin{tabular}{c|cccc}
        \hline\thickhline
        \rowcolor{mygray}
      Method            & PSNR(16 $\times$)        & SSIM(16 $\times$)        & PSNR(32 $\times$)        & SSIM(32 $\times$) \\
      \hline
      pSp               & 18.875            & 0.650              & 18.576            & 0.624 \\
      e4e               & 18.619            & 0.631             & 18.105            & 0.607 \\
      Style Transformer                & 19.152            & 0.676             & 18.830             & 0.646 \\
      \hline
      Ours              & \textbf{20.672}            & \textbf{0.719}             & \textbf{20.147}            & \textbf{0.692} \\
      \hline\thickhline
      \end{tabular}}%
     
    \caption{Quantitative comparison of face super resolution.}
    \label{tab11}%
%yletransformer
  \end{table}%

\paragraph{Results} In Figure~\ref{fig11}, SwinStyleformer achieves satisfactory results on the super resolution. 
We show the results at $32 \times $ and $16 \times $ downsampling ratios, respectively. 
At a ratio of $32 \times $, the SwinStyleformer inversion image is still close to the original image. 
However, other algorithms all generate some differences from the original image, such as the smile details in the first and second rows. 
For a ratio of $16 \times $, SwinStyleformer's inversion results are more realistic compared to others. 
For example, the eye shadow in the third row. The quantitative results of face super resolution in Table~\ref{tab11} show that our method achieves super resolution with sharper output.

\subsection{Efficiency}

\begin{table}[h]
    \centering
    \scriptsize
    \setlength{\abovecaptionskip}{1mm} %调整caption与图的距离
    \setlength{\belowcaptionskip}{-2mm}%调整caption与下文的距离
    \setlength\tabcolsep{2mm}
      \begin{tabular}{cccccccc}
        \hline\thickhline
        \rowcolor{mygray}
      Method  & pSp  & e4e  & st & ReStyle & PTI & StyleRes & Ours \\
      \hline
      time      & 0.088           & 0.089           & 0.063           & 0.365 & 97.942 & 0.125 & 0.086 \\
      \hline\thickhline
    \end{tabular}%
   
    \caption{efficiency comparison (sec). st denotes style transformer. }
    \label{rebuttaltab2}%
%yletransformer
  \end{table}%

In Table~\ref{rebuttaltab2}, we compare the efficiency of our algorithm with other methods. It can be found that our method achieves satisfactory inference times.

\section{Ablation Study}\label{ablation}

\begin{table}[!h]
\centering
\setlength{\abovecaptionskip}{0.1cm} %调整caption与图的距离
\setlength{\belowcaptionskip}{-0.2cm}%调整caption与下文的距离
\resizebox{8.5cm}{!}{
\begin{tabular}{c|ccccc}
\rowcolor{mygray}
\hline\thickhline
& \multicolumn{5}{c}{\text { \textbf{Metric} }}  \\\cline{2-6}
\rowcolor{mygray}
\multirow{-2}{*}{\textbf { Method }} & \text {MSE} & \text{LPIPS} & \text{FID}&\text {PSNR}& \text {SSIM}\\
\hline\hline \text { SwinStyleformer } & 0.032 &  0.159 & 34.218 & 21.274 & 0.617 \\
\text { w/o MSC } & 0.046 & 0.197 & 37.159 & 19.305 & 0.525 \\
\text { w/o DA Loss } & 0.051  &  0.209 & 39.063 & 19.038 & 0.509 \\
\text { w/o inversion discriminator } & 0.038  &  0.184 & 42.624 & 19.668 & 0.541 \\
\text { w/ styleGAN discriminator } & 0.036  &  0.176 & 36.571 & 19.812 & 0.573 \\
\hline\thickhline
\end{tabular}}
\caption{Ablation of multi-scale connections (MSC), distribution alignment (DA) losses, and inversion discriminator.}
\label{appendixtab2}
\end{table}

\paragraph{Core designs} For the ablation of multi-scale connections, we replace them with the pSp pyramid connection approach, and for the ablation of distribution alignment losses and inversion discriminator, we remove them directly, respectively. Meanwhile, we also validate the effectiveness of the inversion discriminator compared to a general discriminator network for the image inversion task. We test it by replacing the inversion discriminator with the discriminator in StyleGAN. Table~\ref{appendixtab2} shows that multi-scale connections, distribution alignment losses, and inversion discriminators are essential for SwinStyleformer. This confirms that minimizing the differences in distribution and multi-scale between Transformer and CNNs is the key to the success of the Transformer in the field of inversion. 

\begin{table}[!h]
\centering
\setlength{\abovecaptionskip}{0.1cm} %调整caption与图的距离
\setlength{\belowcaptionskip}{-0.2cm}%调整caption与下文的距离
\resizebox{8.3cm}{!}{
\begin{tabular}{c|ccccc}
\rowcolor{mygray}
\hline\thickhline
& \multicolumn{5}{c}{\text { \textbf{Metric} }}  \\\cline{2-6}
\rowcolor{mygray}
\multirow{-2}{*}{\textbf { Method }} & \text {MSE} & \text{LPIPS} & \text{FID}&\text {PSNR}& \text {SSIM}\\
\hline\hline \text { w/ ResNet50 } & 0.045 &  0.193 & 41.710 & 19.373 & 0.551 \\
\text {pSp+dilated conv} & 0.036    & 0.168    & 37.574   & 20.920   & 0.592 \\
\text {pSp+Convnext v2} & 0.033    & 0.164    & 35.729   & 21.063   & 0.608 \\
\text { Ours } & 0.032 &  0.159 & 34.218 & 21.274 & 0.617 \\
\hline\thickhline
\end{tabular}}
\caption{Backbone ablation study.}
\label{tab17}
\end{table}

\paragraph{Backbone} In this section, we investigate the effect of convolutional backbone and Transformer backbone on image inversion. When we use ResNet50 as the backbone of our method, the performance of each metric decreases. Meanwhile, We have investigated the impact of dilation convolution and Convnext v2 \cite{woo2023convnext} on the inversion task, we can find that both designs can bring some improvement, but there is still a gap with our method. While Convnext employs a larger kernel and the global design proposed enables the convolutional model to achieve Transformer results. But these designs also affect the local modeling. If the convolutional kernel is too large, it may span larger regions in the image, thus contributing to the model's inability to effectively capture local details. While dilation convolution favors global modeling, the dilation design impairs the continuity of information, which is fatal for pixel-level tasks like inversion. 

\begin{table}[!h]
\centering
\setlength{\abovecaptionskip}{0.1cm} %调整caption与图的距离
\setlength{\belowcaptionskip}{0.3cm}%调整caption与下文的距离
\resizebox{8.3cm}{!}{
\begin{tabular}{c|ccccc}
\rowcolor{mygray}
\hline\thickhline
& \multicolumn{5}{c}{\text { \textbf{Metric} }}  \\\cline{2-6}
\rowcolor{mygray}
\multirow{-2}{*}{\textbf { Method }} & \text {MSE} & \text{LPIPS} & \text{FID}&\text {PSNR}& \text {SSIM}\\
\hline\hline \text { Baseline } & 0.800 &  1.230 & 433.305& 5.234 &0.004\\
\text { w/ W-MSA map2style } & 0.137 &0.412&65.779& 13.286& 0.389 \\
\text { w/ our map2style } & 0.081  &  0.293&51.372& 16.139 & 0.417\\
\hline\thickhline
\end{tabular}}
\caption{Ablation results for W-MSA based on learnable queries.}
\label{appendixtab1}
\end{table}
\vspace{-0.5cm}

\paragraph{Window attention based on learnable queries} A map2style network design based on a window attention mechanism with learnable queries is a prerequisite for a successful SwinStyleformer inversion. We investigate its effects on the face image inversion task. The following ablation studies were all performed on the facial inversion task by default. We employ the pSp coupled with the Swin Transformer backbone as the baseline for window attention based on learnable queries study. The results in Table~\ref{appendixtab1} show that the map2style network with a window attention mechanism based on learnable queries is significant for the success of SwinStyleformer inversion. Figure~\ref{fig4} shows that the map2style network coupled with the window attention mechanism raises artifact issues for the inversion. With the introduction of learnable queries, the artifact problem is successfully solved and thus the inversion metrics are significantly improved.

\begin{figure}[t]
    \setlength{\belowcaptionskip}{0mm}%调整caption与下文的距离
    \centering
    \includegraphics[width=1\linewidth]{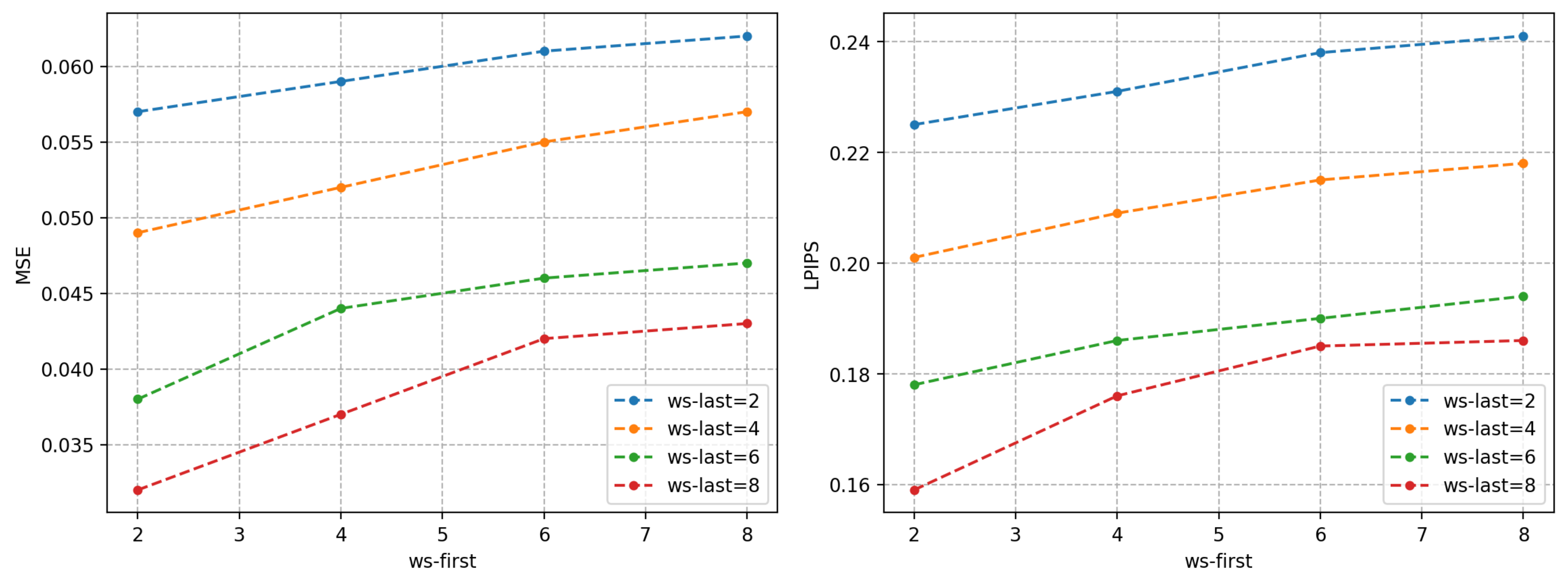}
     \caption{Ablation results for different stage window sizes. $ws-first$ and $ws-last$ indicate the window sizes of the first two phases and the last two phases of Swin Transformer, respectively.}
 \label{appendixfig1}
  \end{figure}

\paragraph{Window size} Regarding the choice of Swin Transformer window size, we have studied it in Figure~\ref{appendixfig1}. The results show that the small size window in the first two stages of Swin Transformer and the large size window in the last two stages are most beneficial for the image inversion of SwinStyleformer. The smaller size windows in the first two stages are designed to capture local details of the image while being more efficient, and the larger size windows in the last two stages retain the Transformer's focus on image structure.

\begin{table}[!h]
\centering
\setlength{\abovecaptionskip}{0.1cm} %调整caption与图的距离
\setlength{\belowcaptionskip}{-0.5cm}%调整caption与下文的距离
\resizebox{8.3cm}{!}{
\begin{tabular}{c|ccccc}
\rowcolor{mygray}
\hline\thickhline
& \multicolumn{5}{c}{\text { \textbf{Metric} }}  \\\cline{2-6}
\rowcolor{mygray}
\multirow{-2}{*}{\textbf { Method }} & \text {MSE} & \text{LPIPS} & \text{FID}&\text {PSNR}& \text {SSIM}\\
\hline\hline \text { all MLP } & 0.291 &  0.526 & 364.069 & 10.847 & 0.067 \\
\text { 2 MLP before 16 length sequences } & 0.045 & 0.201 & 41.197 & 19.529 & 0.573 \\
\text { 1 MLP before 16 length sequences } & 0.036  &  0.170 & 37.105 & 20.972 & 0.606 \\
\text { 1 our module after 16 length sequences } & 0.031  &  0.159 & 34.227 & 21.271 & 0.619 \\
\text { Ours } & 0.032 &  0.159 & 34.218 & 21.274 & 0.617 \\
\hline\thickhline
\end{tabular}}
\caption{Ablation experiments for other components of the map2style network in SwinStyleformer.}
\label{appendixtab3}
\end{table}

\paragraph{Map2style Network} In this section, we investigate the effect of other designs of map2style networks in SwinStyleformer on the inversion results. We focus on the trade-off between the linear layer and the proposed attention module based on learnable queries. For the above ablation study, we designed four ablation settings separately.

\begin{figure}[t]
    \setlength{\belowcaptionskip}{-5mm}%调整caption与下文的距离
    \centering
    \includegraphics[width=1\linewidth]{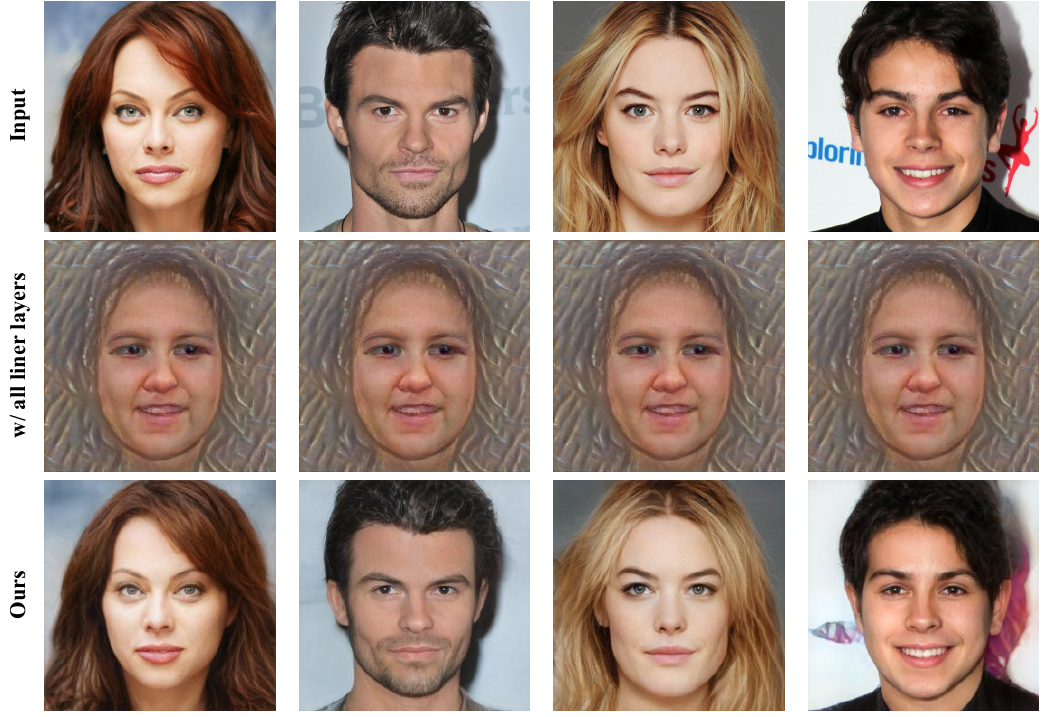}
     \caption{Visualization results of a linear layer number ablation study for map2style networks.}
 \label{appendixfig2}
  \end{figure}

We first study the replacement of all modules in the map2style network with linear layers. The results are shown in Table~\ref{appendixtab3} and Figure~\ref{appendixfig2}, where we find that the map2style network with all linear layers is not sufficient to reconstruct the input image. Next, we study the effect of adding $1$ or $2$ linear layers before a sequence of tokens of length $16$, respectively. In addition, we study the effect of replacing $1$ linear layer after a token of length $16$ with a Transformer module based on learnable queries. Experiments show that the use of linear layers for token sequences up to length $16$ affects the inversion results to some extent. The use of the Transformer module based on learnable queries for token sequences after length $16$ basically does not change the experiment results. Therefore, considering the trade-off between computational cost and inversion accuracy, we finally choose to use the Transformer module based on learnable queries for token sequences up to length $16$ and use linear layer processing for token sequences after $16$.

\begin{table}[!h]
\centering
\setlength{\abovecaptionskip}{0.1cm} %调整caption与图的距离
\setlength{\belowcaptionskip}{-0.1cm}%调整caption与下文的距离
\resizebox{8.3cm}{!}{
\begin{tabular}{c|ccccc}
\rowcolor{mygray}
\hline\thickhline
& \multicolumn{5}{c}{\text { \textbf{Metric} }}  \\\cline{2-6}
\rowcolor{mygray}
\multirow{-2}{*}{\textbf { Method }} & \text {MSE} & \text{LPIPS} & \text{FID}&\text {PSNR}& \text {SSIM}\\
\hline\hline \text { w/o aps } & 0.037 &  0.175 & 37.834 & 20.728 & 0.607 \\
\text { w/o normalization } & 0.033 & 0.162 & 34.735 & 20.910 & 0.616 \\
\text { w/ conv } & 0.039  &  0.182 & 38.294 & 20.136 & 0.592 \\
\text { Ours } & 0.032 &  0.159 & 34.218 & 21.274 & 0.617 \\
\hline\thickhline
\end{tabular}}
\caption{Ablation experimental results of the SwinStyleformer upsampling module, the absolute position encoding (aps), and the normalization in the upsampling module. conv denotes the upsampling module with convolutional layers.}
\label{appendixtab4}
\end{table}

\paragraph{Upsample Module}\label{upsample} For the upsampling module in the multi-scale connections, we ablate the Patch Expanding strategy, the absolute position encoding, and the normalization operation in it. Regarding the Patch Expanding strategy, we use $1\times1$ convolution for its ablation. Table~\ref{appendixtab4} shows that the multi-scale connection using convolution can affect the inversion to some extent. The inversion effect of SwinStyleformer deteriorates when either the normalization operation or the absolute position encoding is removed. This illustrates the role of Patch Expanding strategy, normalization operations, and absolute position encoding in SwinStyleformer.

\section{Conclusion}

In this paper, we successfully solve the Transformer inversion failure issue and thus propose the first pure Transformer structure image inversion network SwinStyleformer. 
Compared with previous image inversion frameworks, SwinStyleformer is designed to better handle the global structure while preserving local details with its core design such as multi-scale connections, Transformer module based on learnable queries, distributed alignment losses, and inversion discriminators. 
As a result, SwinStyleformer demonstrates SOTA performance in image inversion and several other vision tasks related to it.

\section*{Acknowledgments}
This work was supported by Public-welfare Technology Application Research of Zhejiang Province in China under Grant LGG22F020032, Basic Public-Welfare Research Project of Wenzhou in China under Grant G2023093,  and Key Research and Development Project of Zhejiang Province in China under Grant 2021C03137.

\clearpage
\appendix

\section{Datasets Details}

For image inversion and image inversion for specific details in the facial domain, we select FFHQ~\cite{karras2019style} containing $70K$ images as the training set and randomly select $6K$ images from the CelebA-HQ~\cite{karras2017progressive} dataset of $30K$ images as the test set according to the setting of pSp~\cite{richardson2021encoding}. The image editing, image editing for specific details, and style mixing tasks are performed on the test set of CelebA-HQ. For super resolution task, here we choose CelebA-HQ dataset for training and testing. Regarding the semantic segmentation map to face task, we perform the inversion on the CelebAMask-HQ dataset.

For the image inversion in the animal domain, we adopt the AFHQ~\cite{choi2020stargan} Wild dataset for training and testing. The inversion task for the church domain is carried on the LSUN~\cite{yu2015lsun} Church dataset.

\section{Total Loss}

In addition to the distribution alignment loss and the inversion discriminator. We follow the losses of pSp to constrain inversion training, including $L2$ loss, LPIPS \cite{zhang2018unreasonable} loss for perceptual similarity, and ID \cite{richardson2021encoding} loss for preserving input identity.
\begin{equation}
    \begin{split}
&L({\rm{x}}) = {\lambda _1}{L_2}({\rm{x}}) + {\lambda _2}{L_{{\rm{LPIPS}}}}({\rm{x}}) \\
&+ {\lambda _3}{L_{{\rm{ID}}}}({\rm{x}}) + {\lambda _4}{L_{{\rm{DA}}}}({\rm{x}}) + {\lambda _5}{L_{{\rm{adv}}}}({\rm{x}}),
\end{split}
\label{eq2}
\end{equation}
where $L_{DA}(.)$ is the distribution alignment loss and $L_{adv}(.)$ is the adversarial loss. 
$\lambda_1$, $\lambda_2$, $\lambda_3$, $\lambda_4$, $\lambda_5$ are the hyperparameters. 
These loss functions help SwinStyleformer to achieve effective inversion while reducing the distribution difference between the latent code and the style vector.

\section{Implementation Details}

In this section, we describe the specific implementation details. For the SwinStyleformer backbone, we choose the tiny Swin Transformer~\cite{liu2021swin} (Swin-T) with a patch size default of $4$. Regarding the first two stages of Swin-T, we default the window size to $2$, and the last two stages still keep the window size of $8$. The rest of the settings are consistent with the Swin-T. In addition, the window attention based on learnable queries in map2style employs a window of size $2$ to strengthen the local modeling. According to the setting of~\cite{tov2021designing,hu2022style,wang2022high,richardson2021encoding}, SwinStyleformer still utilizes the fixed pre-trained StyleGAN2~\cite{karras2019style,karras2020analyzing} generator to complete the inversion and related tasks. The rest of the architectural details are described in the main text.

To reduce computational costs, the SwinStyleformer input image resolution and the StyleGAN2 output image resolution are maintained at $256\times256$ in all tasks. Concerning the choice of hyperparameters, we keep $1$, $0.8$, and $0.1$ as settings in pSp~\cite{richardson2021encoding} for the hyperparameters $\lambda_1,\lambda_2,\lambda_3$ of mean square error, LPIPS~\cite{zhang2018unreasonable} loss, and ID~\cite{richardson2021encoding} loss. For the corresponding hyperparameters $\lambda_4,\lambda_5$ for proposed distribution alignment loss and the adversarial loss provided by the inversion discriminator, we default to $0.1$ and $1e-4$ to prevent excessively affecting the latent code distribution. All tasks are performed in the PyTorch framework with only a single NVIDIA RTX 3090.

Regarding the training settings of the SwinStyleformer’s inversion discriminator, we apply an unbalanced data augmentation strategy of~\cite{chen2021mocov3} to the input image as well as the inversion image to enlarge the discrepancy. The inversion discriminator and SwinStyleformer training both use a learning rate with a $1e-4$ Ranger~\cite{Ranger} optimizer. Due to the limitation of computational resources, SwinStyleformer can only use a batch size of $6$. But the rest of SwinStyleformer's training settings are consistent with pSp. Besides, the rest of the training settings in the ablation experiment remained the same as the above training settings except for the ablation factor.

Regarding the evaluation metrics, apart from the MSE and LPIPS metrics for similarity evaluation, we also selected FID, PSNR, and SSIM as the evaluation metrics for SwinStyleformer in order to make a more comprehensive comparison.

\begin{figure*}[!t]
    \centering
    \setlength{\abovecaptionskip}{-0.1mm} %调整caption与图的距离
    \setlength{\belowcaptionskip}{-3mm}%调整caption与下文的距离
    \includegraphics[width=0.93\linewidth]{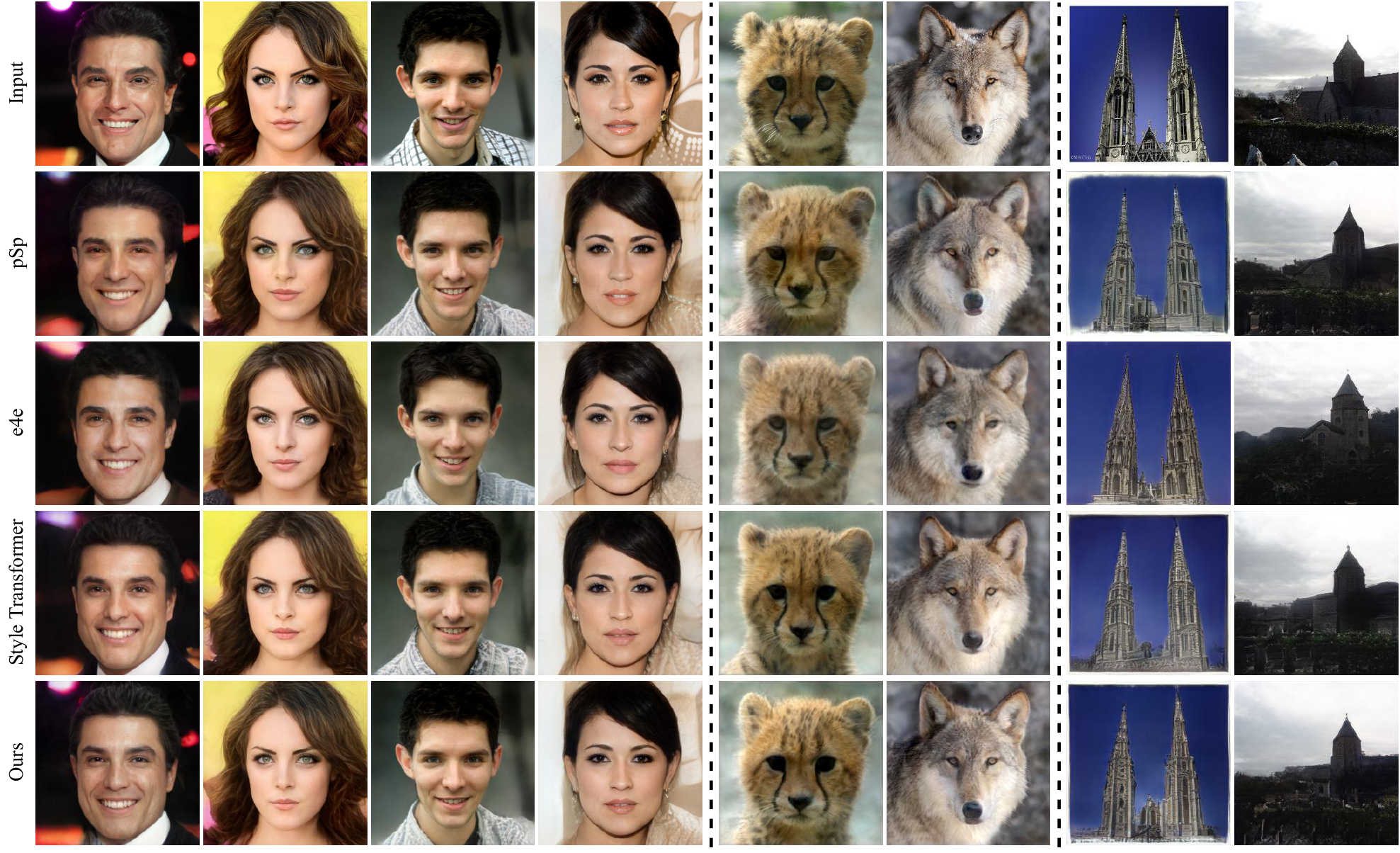}
     \caption{More inversion comparison results on different domains.}
     \label{appendixfig3}
   \end{figure*}

\begin{figure*}[!t]
    \centering
    \setlength{\abovecaptionskip}{-0.1mm} %调整caption与图的距离
    \setlength{\belowcaptionskip}{-5mm}%调整caption与下文的距离
    \includegraphics[width=0.92\linewidth]{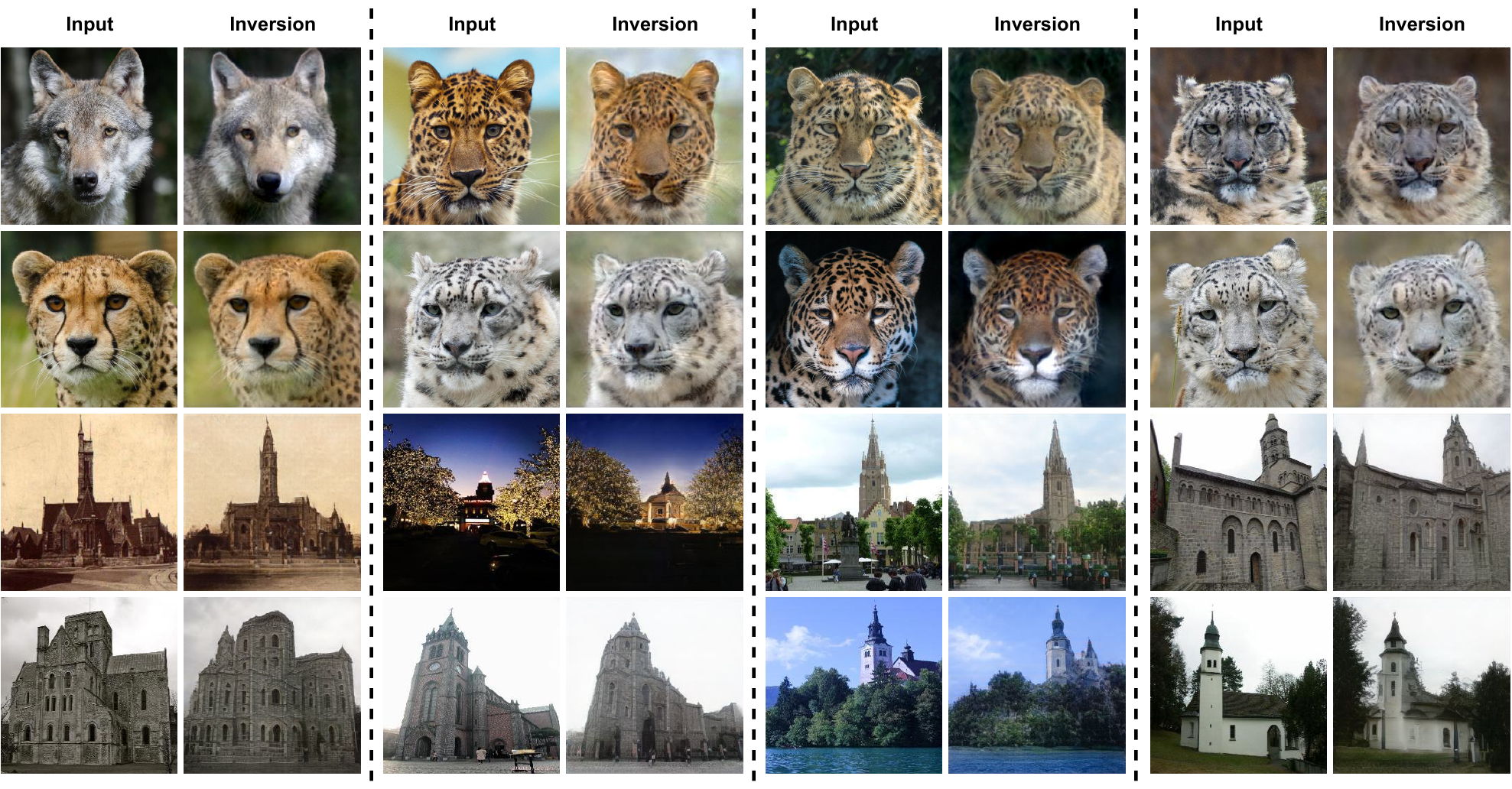}
     \caption{More inversion results on animal and church domains.}
     \label{appendixfig5}
   \end{figure*}

\begin{figure*}[!t]
    \centering
    \setlength{\belowcaptionskip}{-5mm}%调整caption与下文的距离
    \includegraphics[width=0.93\linewidth]{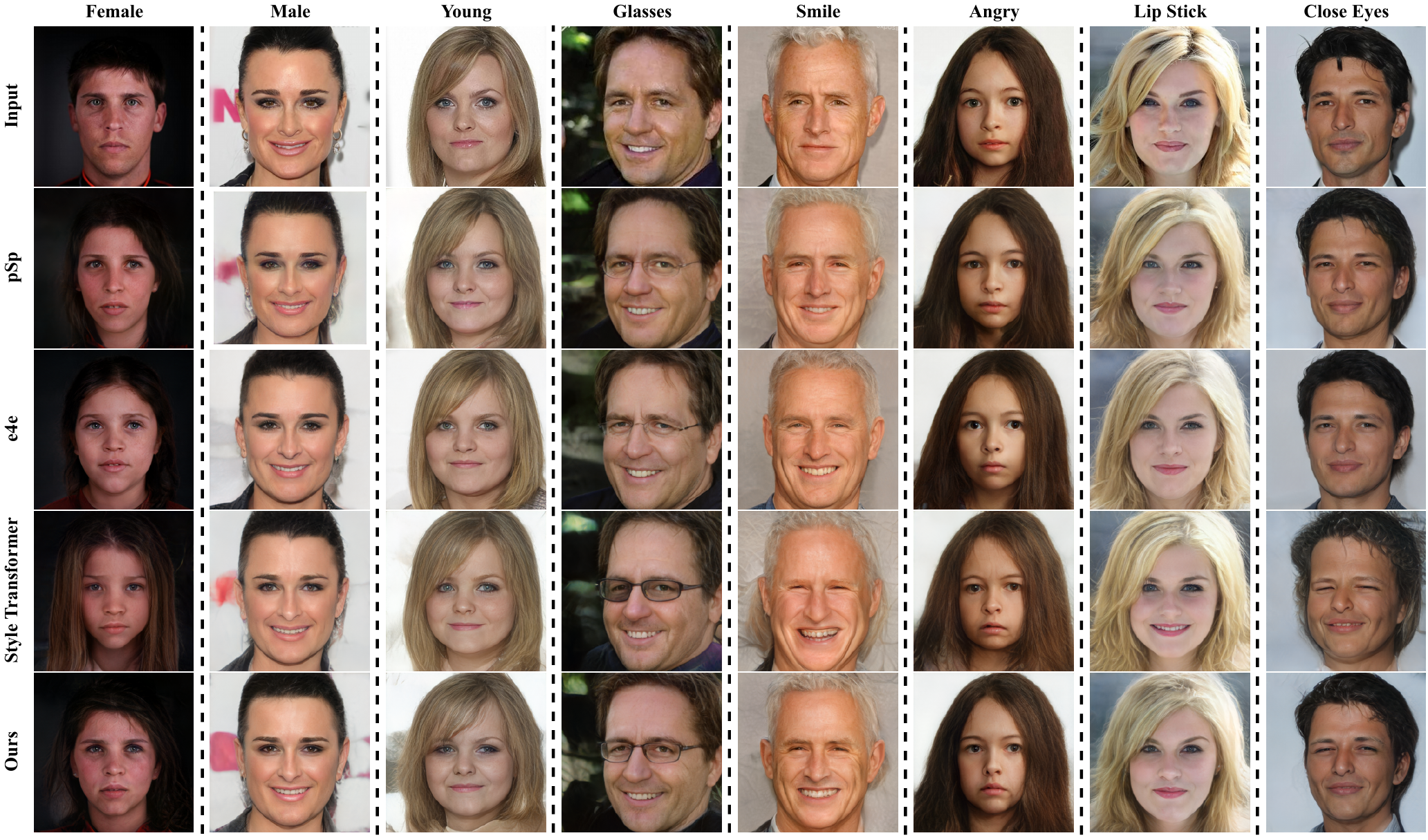}
     \caption{More qualitative comparison results for image editing.}
     \label{appendixfig8}
   \end{figure*}

\begin{figure}[t]
    \setlength{\belowcaptionskip}{-5mm}%调整caption与下文的距离
    \centering
    \includegraphics[width=1\linewidth]{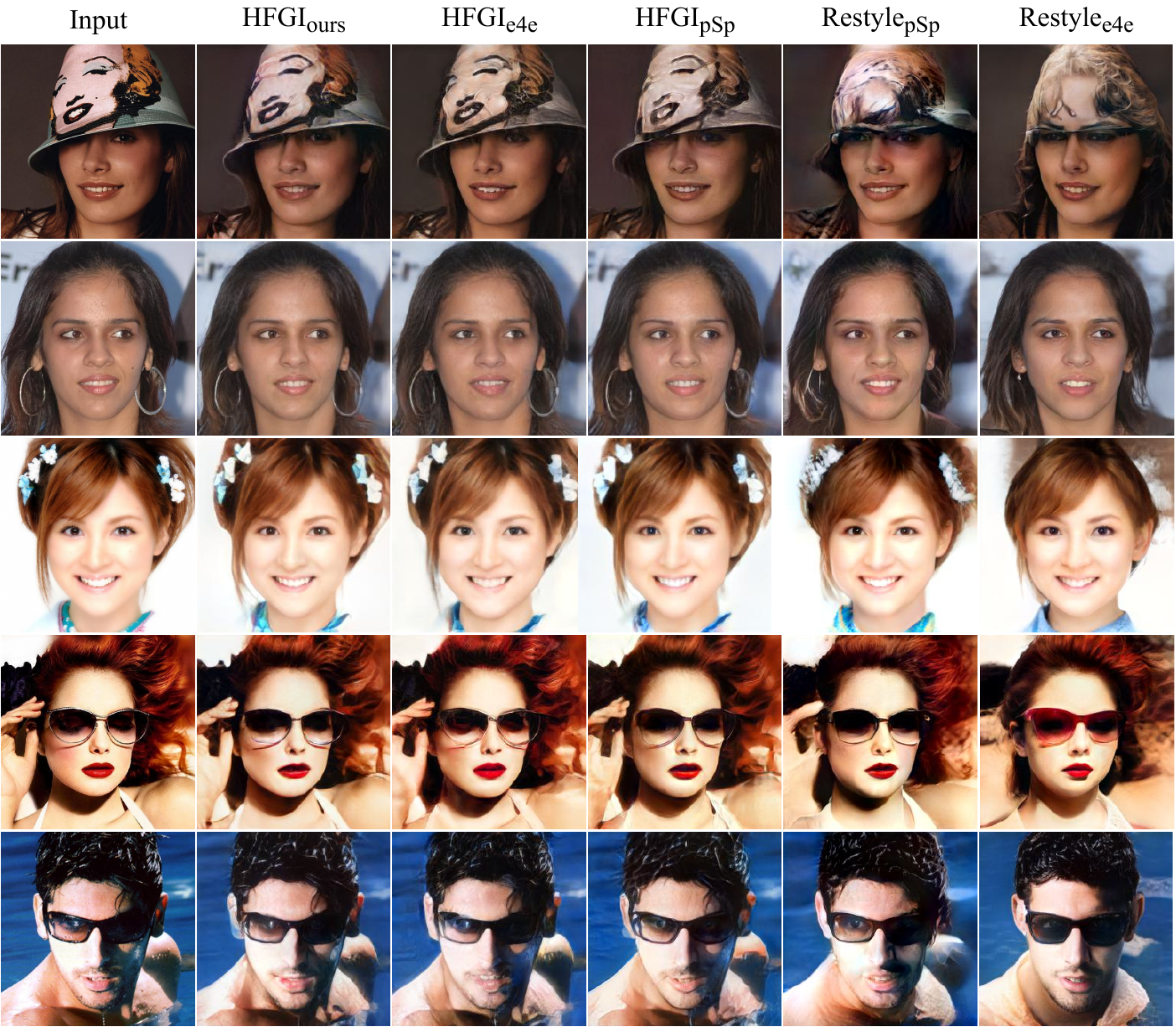}
     \caption{More comparison results between SwinStyleformer and baselines for image inversion for specific details.}
 \label{appendixfig6}
  \end{figure}

\begin{figure*}[!t]
    \centering
    \setlength{\belowcaptionskip}{-5mm}%调整caption与下文的距离
    \includegraphics[width=0.93\linewidth]{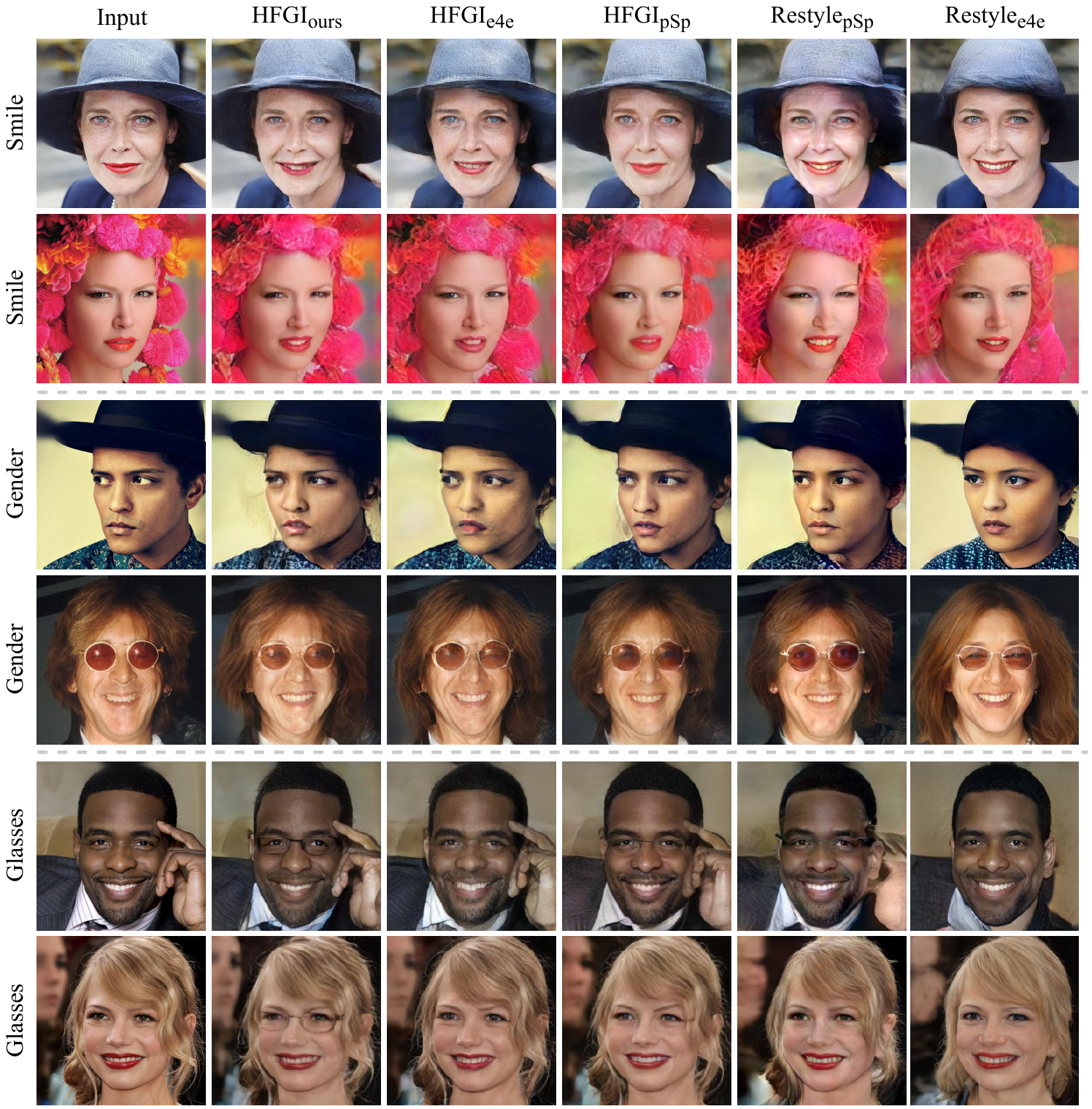}
     \caption{More qualitative comparison results of image editing for specific details.}
     \label{appendixfig9}
   \end{figure*}

\begin{figure*}[!t]
    \centering
    \setlength{\belowcaptionskip}{-5mm}%调整caption与下文的距离
    \includegraphics[width=0.93\linewidth]{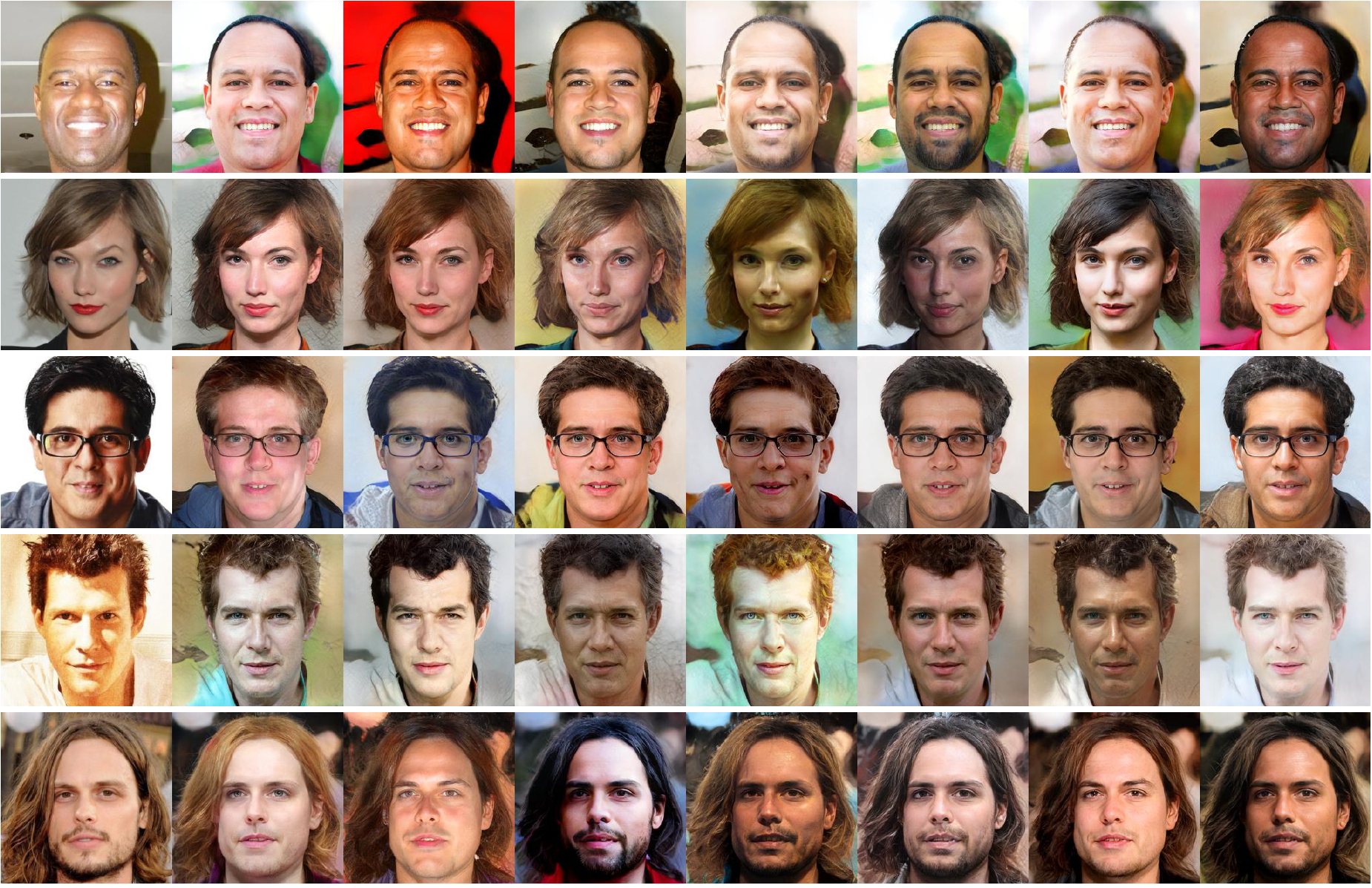}
     \caption{Style mixing results of SwinStyleformer on CelebA-HQ.}
     \label{appendixfig10}
   \end{figure*}

\begin{figure}[t]
    \setlength{\belowcaptionskip}{-5mm}%调整caption与下文的距离
    \centering
    \includegraphics[width=1\linewidth]{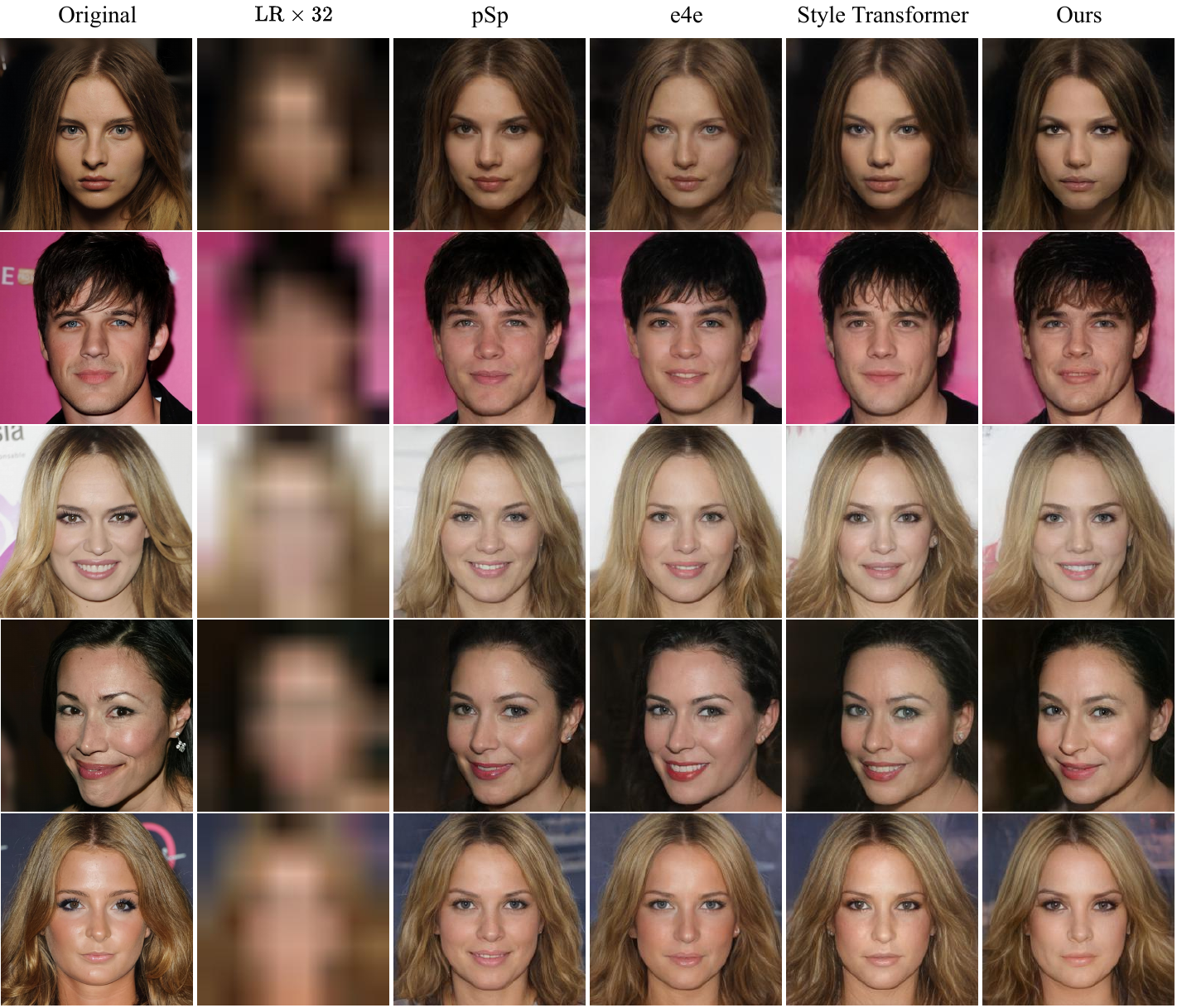}
     \caption{More qualitative comparison of super-resolution at $32\times$ downsampling ratio.}
 \label{appendixfig11}
  \end{figure}

\begin{figure}[t]
    \setlength{\belowcaptionskip}{-5mm}%调整caption与下文的距离
    \centering
    \includegraphics[width=1\linewidth]{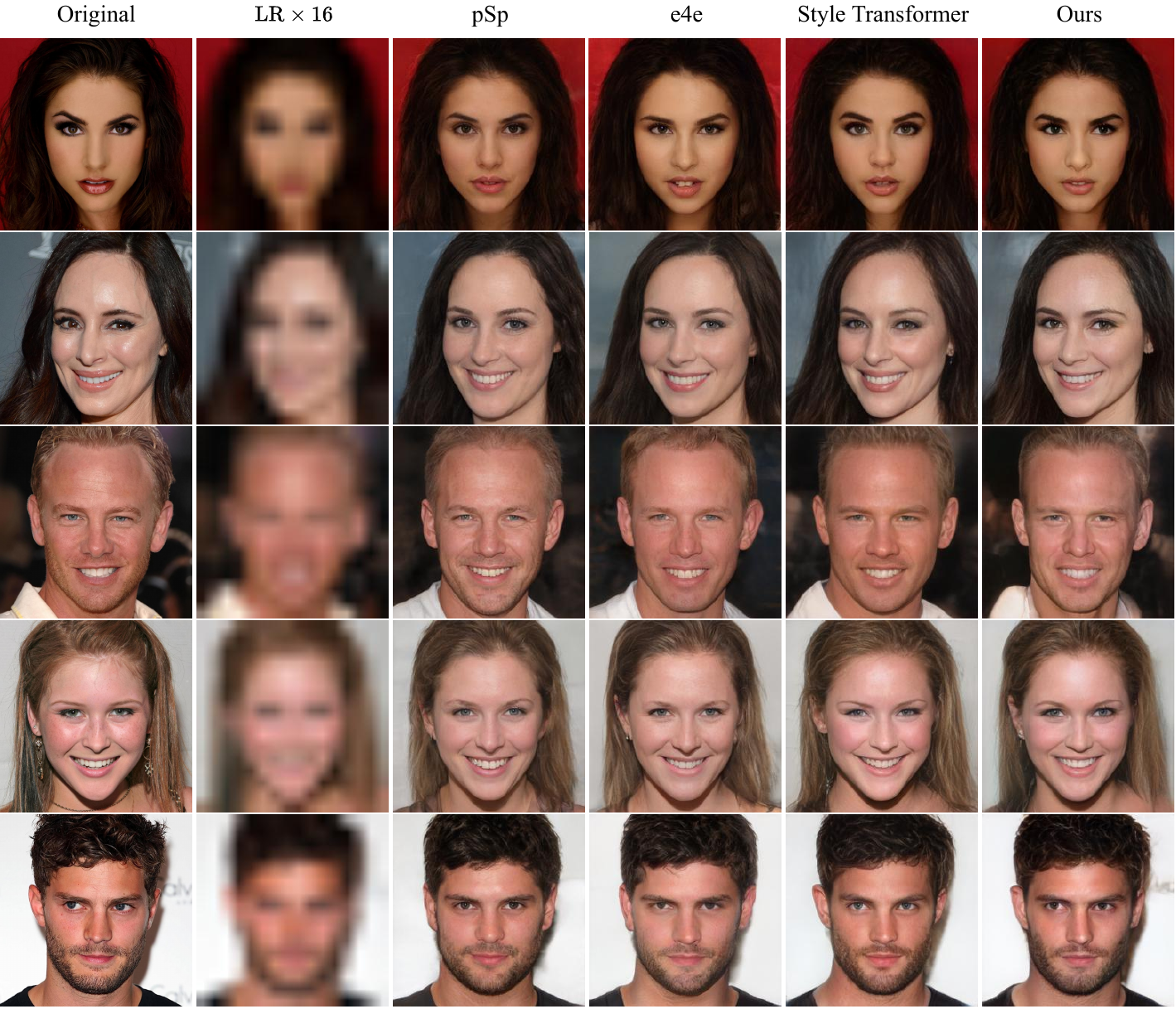}
     \caption{More qualitative comparison of super-resolution at $16\times$ downsampling ratio.}
 \label{appendixfig12}
  \end{figure}

\begin{figure*}[!t]
    \centering
    \setlength{\belowcaptionskip}{-5mm}%调整caption与下文的距离
    \includegraphics[width=0.93\linewidth]{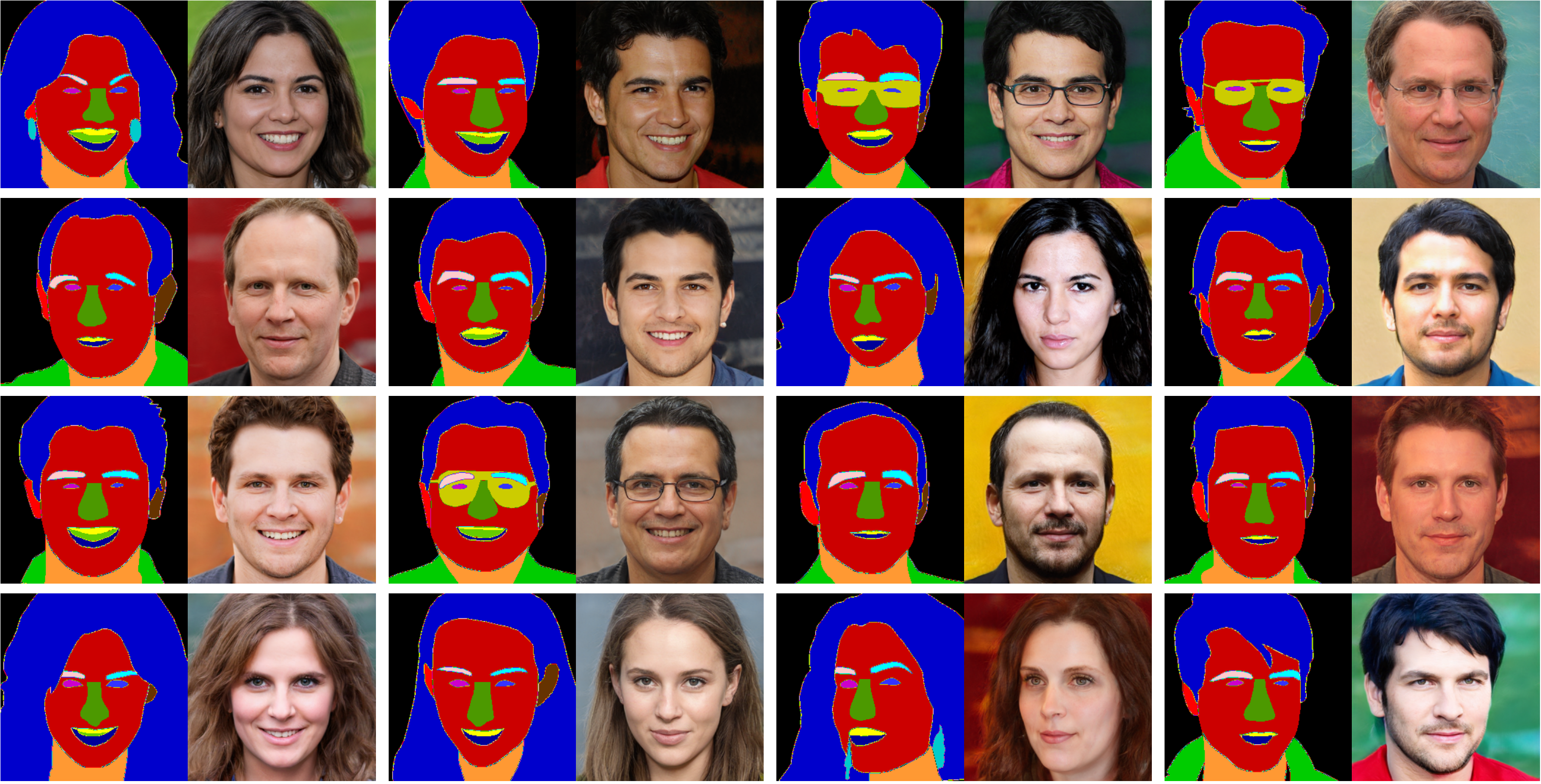}
     \caption{More results for SwinStyleformer's semantic segmentation map to face.}
     \label{appendixfig14}
   \end{figure*}

\section{Experiment Details and Results}

\subsection{Image Inversion}

For SwinStyleformer, we use the implementation details in Appendix \textbf{C}. For the baselines of the comparisons, we followed the official code to train and test in a batch size of $8$ on the datasets corresponding to the three domains.

\paragraph{Results.} Figure~\ref{appendixfig3} shows more comparison results of SwinStyleformer with the baseline in the three domains. It can be noticed that our method shows more details compared to the baseline. For example, the smile details in the images in the first and third columns. At the same time, the baselines in the fourth column all invert the non-existent hair for the input image, while SwinStyletransformer makes an accurate judgment. Figure~\ref{appendixfig4} and Figure~\ref{appendixfig5} show more facial inversion results, animal inversion results, and church inversion results, respectively.

\subsection{Image Inversion for Specific Details}

Regarding the image inversion for details, SwinStyleformer, as the backbone of HFGI~\cite{wang2022high}, conducts inversion training $100K$ epochs according to a batch size of $6$. For the loss function, we only adopt the loss function used in the main text of HFGI for training. About the rest of the baselines paired with HFGI, they all keep the same settings except that they use a batch size of 8. Other baselines are trained according to the official settings. Next, we show more comparison results and image inversion results for specific details in Figure~\ref{appendixfig6} and Figure~\ref{appendixfig7}.

\paragraph{Results.} Figure~\ref{appendixfig6} shows that our method is more accurate for the inversion of specific details. For the hat in the first row of inversion results, the HFGI with both e4e and pSp baselines produces some degree of chromatic aberration compared to the original image while the Restyle~\cite{alaluf2021restyle} baseline failed for its inversion. In contrast, the hat color of the SwinStyleformer inversion result is closer to the input image. HFGI with SwinStyleformer successfully generated the earrings in the second row of images, and the earrings in the other baseline inversion results are less regular in shape. In the fourth row of inversion results, the hand inversion of our method is more realistic.

\subsection{Image Editing}\label{sec:edit}

According to the editing directions given by InterfaceGAN~\cite{shen2020interpreting}, we edit the latent space of SwinStyleformer and other baselines trained by inversion. We selected the editing direction of smiling, female, male, young, glasses, angry, lipstick, and closed eyes. More comparison results and image editing results are given in Figure~\ref{appendixfig8} and Figure~\ref{appendixfig15}.

\paragraph{Results.} Compared to other baselines, SwinStyleformer achieves more significant editing while retaining similarity to the input image. For the female direction edits, all baseline performances are significant. However, they all have some differences with the input images. In contrast, SwinStyleformer's editing results are closer to the input image. For edits in other directions, the editing results of our method and Style Transformer are more obvious compared to other baselines for the editing same weight. However, there are some minor flaws in the Style Transformer editing results such as chromatic aberrations on angry edits, artifacts on smile and closed-eye edits, etc.

\subsection{Image Editing for Specific Details}

Consistent with the setting in section~\ref{sec:edit}, we choose editing directions such as smile, gender, etc. to show the effect of image editing for specific details. We edit the image directly according to the official code of HFGI and Restyle.

\paragraph{Results.} Figure~\ref{appendixfig9} shows more comparison results of HFGI coupled with SwinStyleformer with other baselines. As can be seen, SwinStyleformer successfully achieves image editing while preserving image details. For example, in the first row of edited results, all baseline hats produce some degree of artifacts. In contrast, the hats in the SwinStyleformer's edit results are more realistic. For the gender edit in the fourth row, we have more significant results compared to the other baselines. Although Restyle successfully implements gender editing, it also produces relatively large differences from the input image. In terms of glasses editing, compared to other baselines, SwinStyleformer achieves a more obvious edit while retaining more detail (e.g., hands, background, etc.).

\subsection{Style Mixing}\label{sec:mix}

In this section, we also test the performance of SwinStyleformer on StyleGAN's classical style mixing task. We interpolate layers $8,9,10,11,12,13$ of the SwinStyleformer latent code. Regarding the interpolation vector, we use the style vector obtained by StyleGAN through random noise.

\paragraph{Results.} Figure~\ref{appendixfig10} shows the results of SwinStyleformer style mixing. The leftmost is the input image and the rest are interpolated images. We can see that SwinStyleformer performs well in the style mixing task. Thus, it is further shown that the latent code obtained by SwinStyleformer inversion successfully approximates the style vector distribution of StyleGAN.

\subsection{Super Resolution}

Regarding the super resolution training, we randomly downsample the input images at $1,2,4,8,16,32$ factors according to the pSp setting in the training phase. The target image remains unchanged. In addition, we use the regularization loss proposed by pSp during the super-resolution training to encourage the encoder output to be closer to the average style vector. The loss weights default to $0.005$. Finally, we show more super resolution results of SwinStyleformer at $32\times$ downsampling ratio and $16\times$ downsampling ratio in Figure~\ref{appendixfig11} and Figure~\ref{appendixfig12}.

\paragraph{Results.} In Figure~\ref{appendixfig11} and Figure~\ref{appendixfig12}, SwinStyleformer recovers more detail in the super resolution task compared to the other baselines. For example, in the third row of Figure~\ref{appendixfig11}, the smile detail of the SwinStyleformer's super resolution results is closer to the original image. For the details of the lips in the fourth row, all baseline's super resolution results show relatively large differences compared to the original image. However, the lips details recovered by SwinStyleformer are more similar to the original image. For the eye detail in the first row, the face size in the third row, and the eye orientation in the fifth row in Figure~\ref{appendixfig12}, our method is closer to the original image compared to the baselines.

\subsection{Face from Segmentation Map}

\begin{figure*}[t]
    \setlength{\belowcaptionskip}{-5mm}%调整caption与下文的距离
    \centering
    \includegraphics[width=1\linewidth]{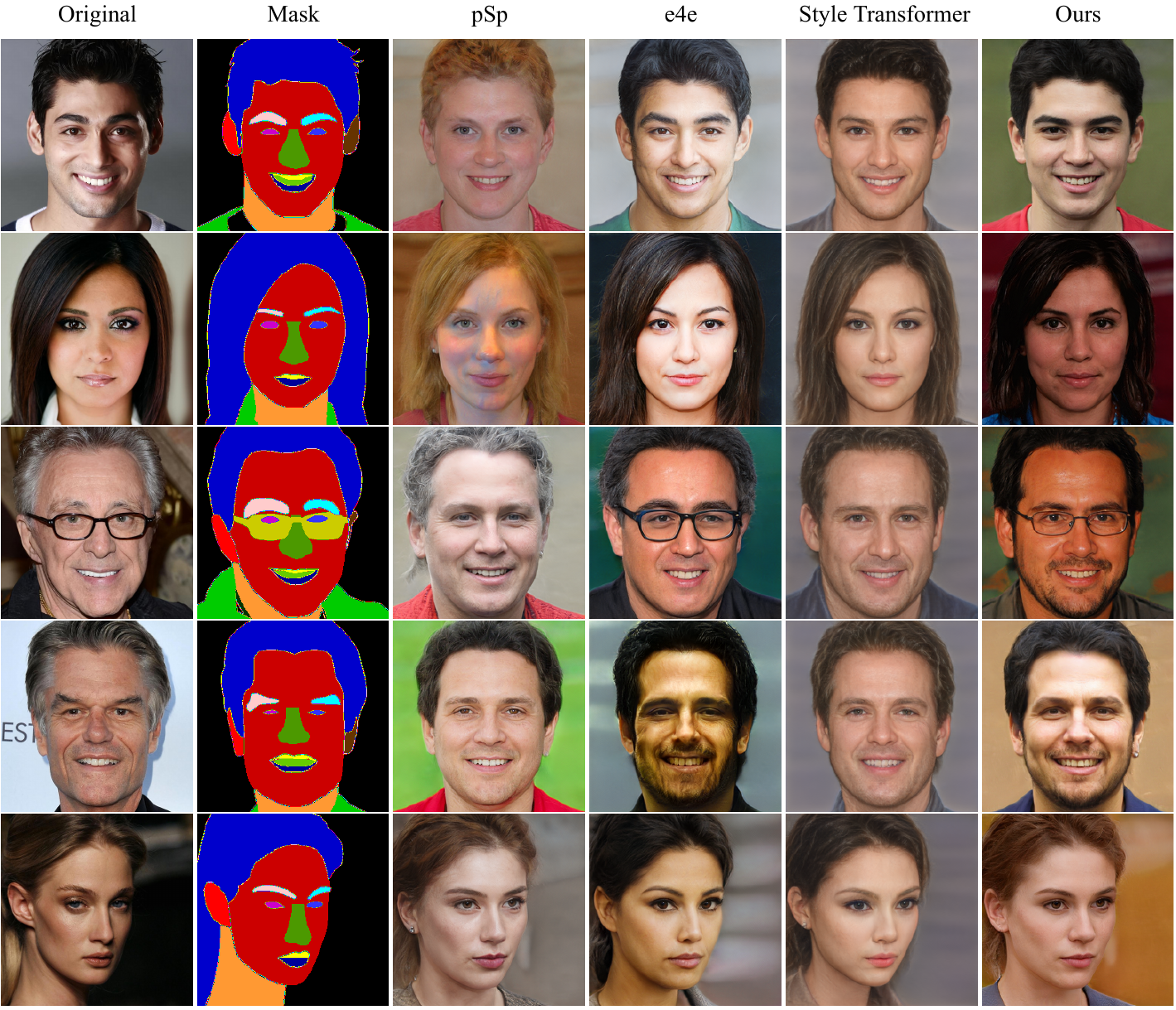}
     \caption{More semantic segmentation maps to face comparison results.}
 \label{appendixfig13}
  \end{figure*}

We trained and tested the semantic segmentation map to face for SwinStyleformer with the CelebAMask-HQ dataset. We first process the semantic segmentation map in CelebAMask-HQ in the same way as pSp. Then, SwinStyleformer generates a diverse output for the semantic segmentation map based on style mixing. The settings for style mixing follow those in section~\ref{sec:mix}. Since Style Transformer using style mixing can seriously affect the quality of the semantic segmentation map to the face, we do not perform style mixing for Style Transformer. Otherwise, all baselines are trained and tested according to the same settings.

Results. Figure~\ref{appendixfig13} shows more comparison results of semantic segmentation map to face. It can be found that our method is more accurate for the inversion of the semantic segmentation map. For example, SwinStyleformer and e4e accurately inverse the glasses in the semantic segmentation map of the third row. Compared to e4e, our method provides more facial details. For the inversion of the semantic segmentation map in the fourth row, SwinStyleformer successfully inverts the shape of the brow corner that other baselines failed to invert.

\section{More Results}

We provide more editing results and inversion results for specific details in other domains in this section. Figure~\ref{appendixfig15} shows more results of the image editing with SwinStyleformer. Figure~\ref{appendixfig16} shows the inversion results of SwinStyleformer with HFGI on more domains.

\clearpage

 \begin{figure*}[!t]
    \centering
    \setlength{\belowcaptionskip}{-5mm}%调整caption与下文的距离
    \includegraphics[width=0.93\linewidth]{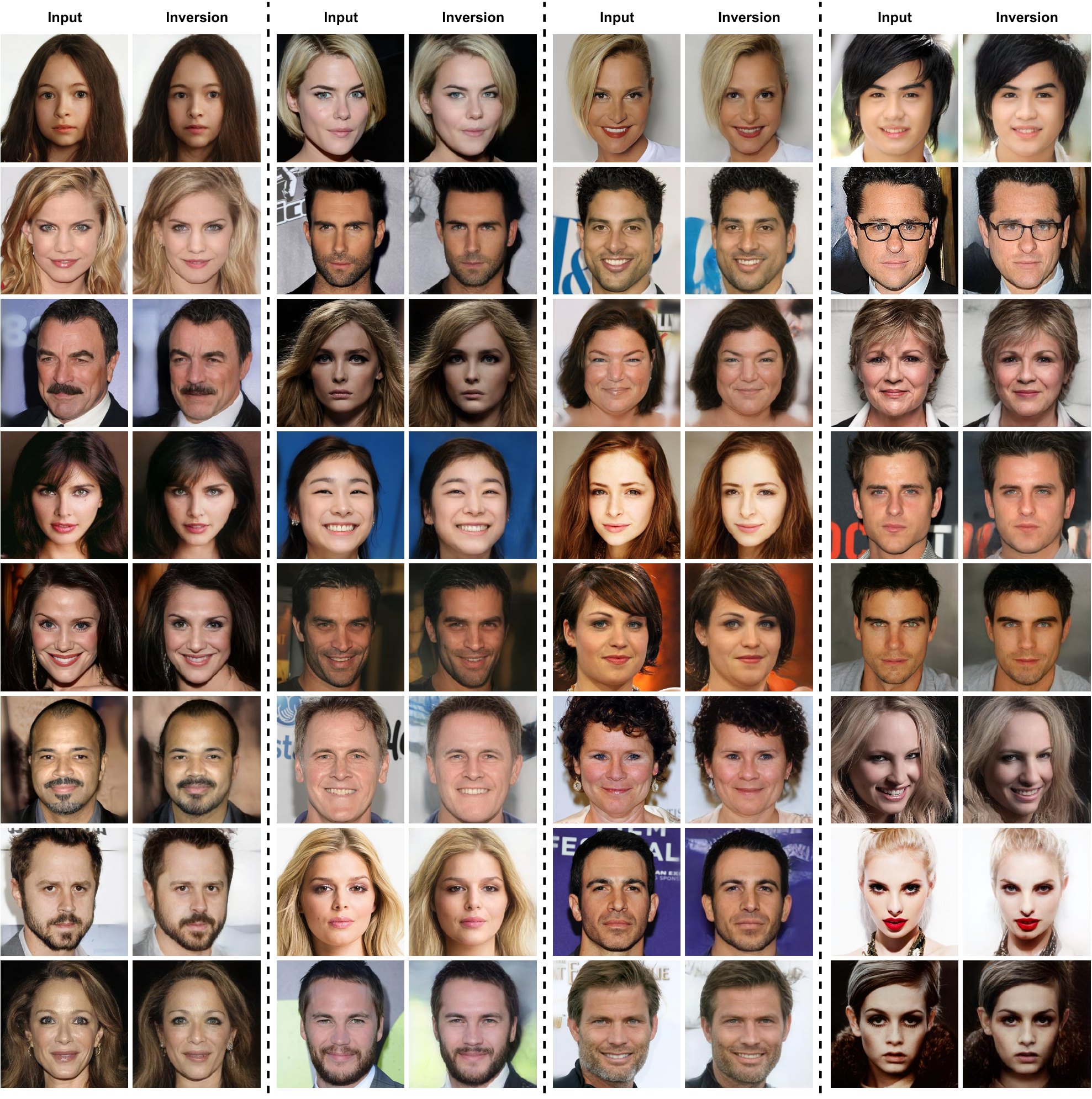}
     \caption{More facial inversion results for SwinStyleformer.}
     \label{appendixfig4}
   \end{figure*}

\begin{figure*}[!t]
    \centering
    \setlength{\belowcaptionskip}{-5mm}%调整caption与下文的距离
    \includegraphics[width=0.93\linewidth]{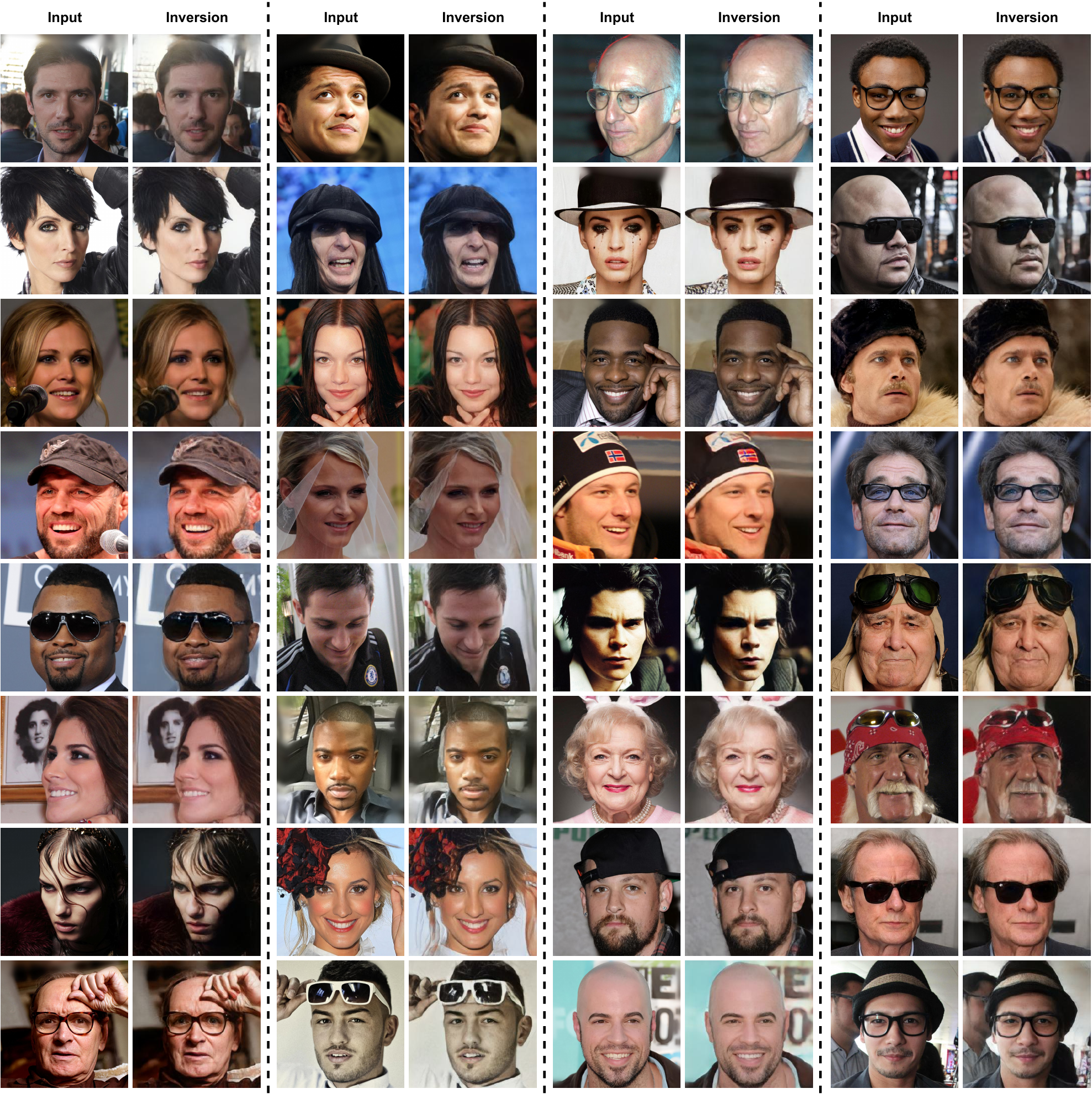}
     \caption{More inversion results for specific details.}
     \label{appendixfig7}
   \end{figure*}

\begin{figure*}[!t]
    \centering
    \setlength{\belowcaptionskip}{-5mm}%调整caption与下文的距离
    \includegraphics[width=0.93\linewidth]{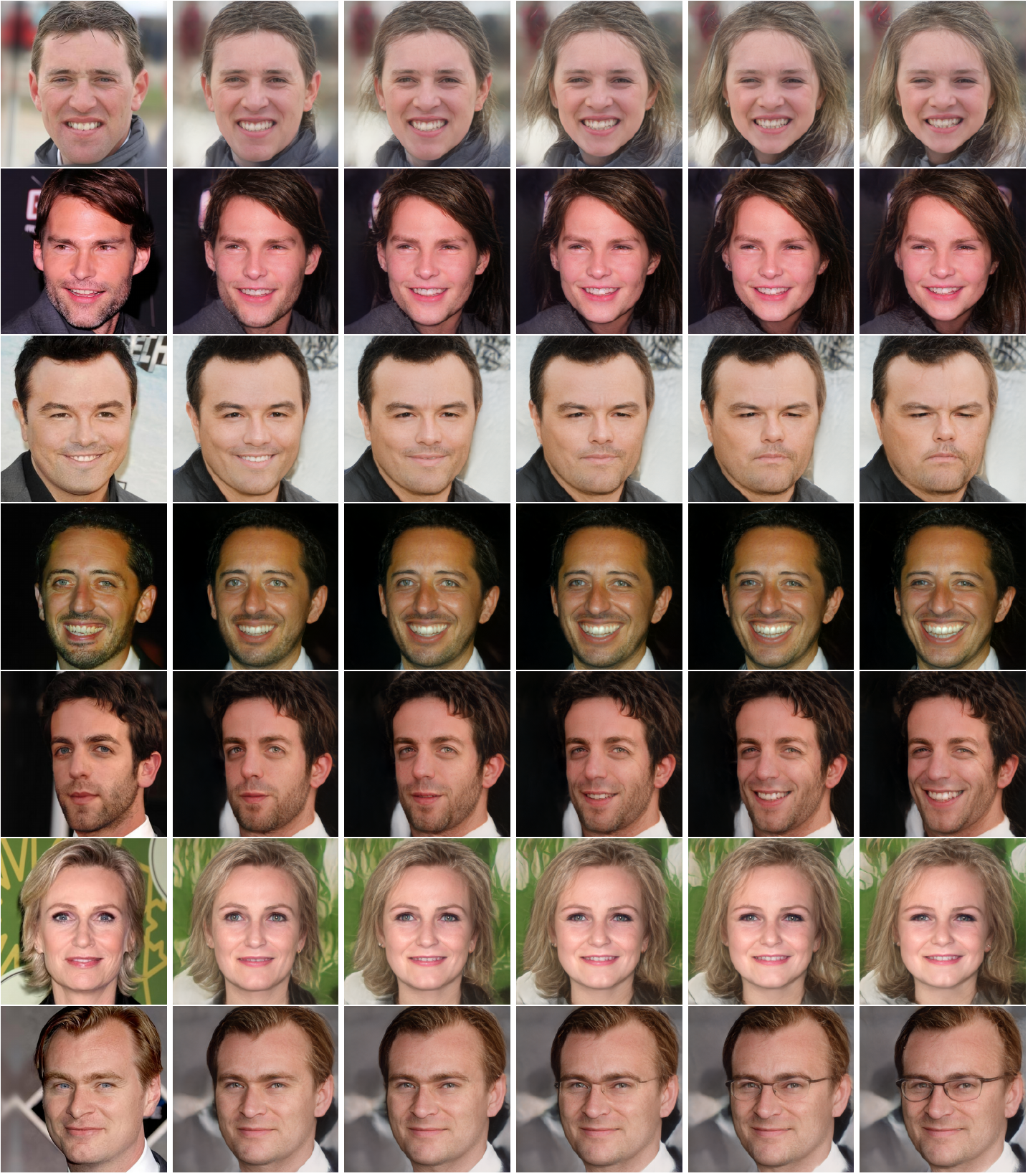}
     \caption{More facial editing results for SwinStyleformer.}
     \label{appendixfig15}
   \end{figure*}

\begin{figure*}[!t]
    \centering
    \setlength{\belowcaptionskip}{-5mm}%调整caption与下文的距离
    \includegraphics[width=0.93\linewidth]{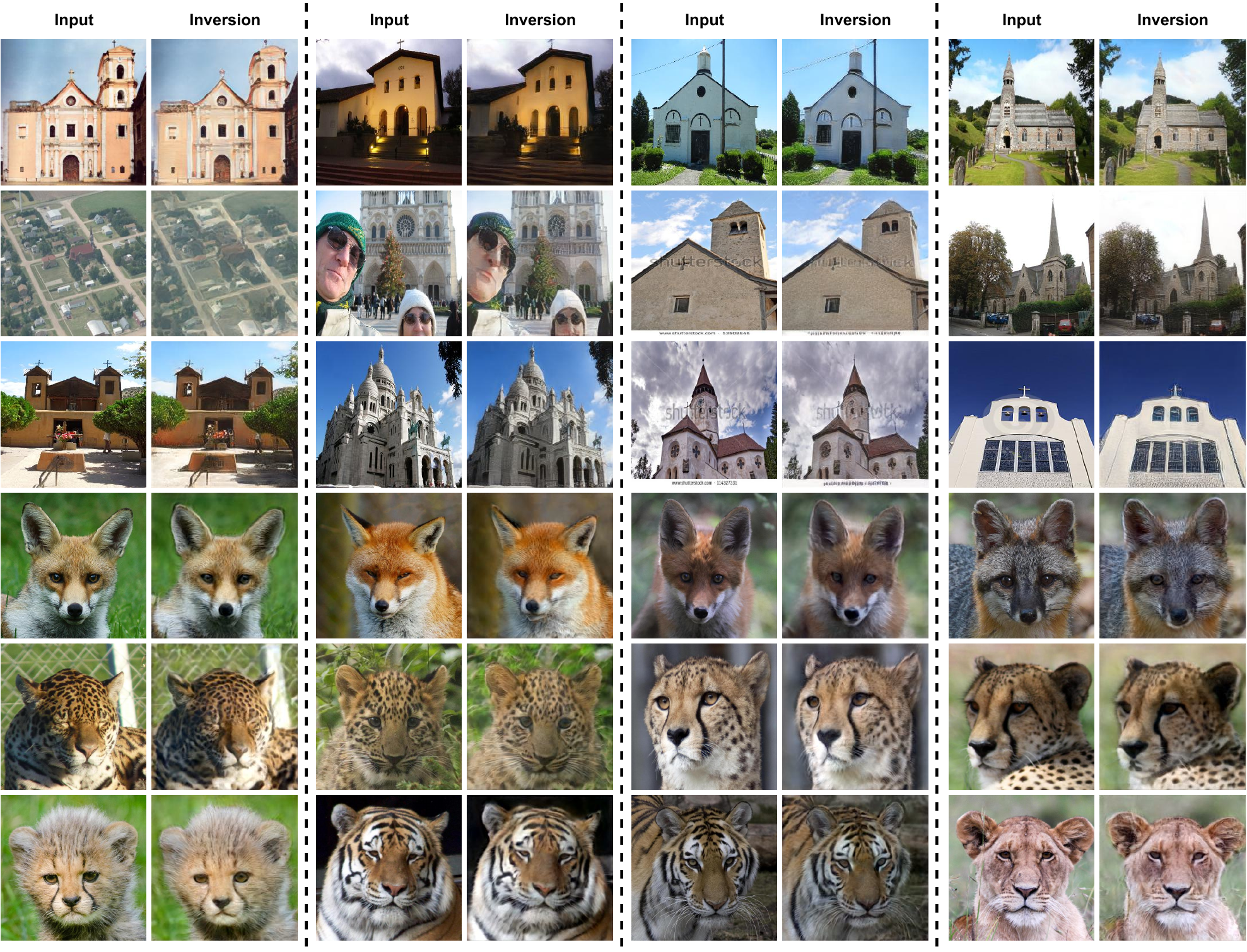}
     \caption{Inversion results for specific details on different domains.}
     \label{appendixfig16}
   \end{figure*}

\clearpage

 \bibliographystyle{ieeetrans}
 \bibliography{main}

\end{document}